\documentclass[journal]{IEEEtran}

\usepackage[cmex10]{amsmath}
\usepackage{amssymb}
\usepackage{algorithm}
\usepackage{algorithmic}
\usepackage{graphicx}
\usepackage{array}
\usepackage{amsmath}
\usepackage{subfig}
\usepackage{multirow}   %????????multirow????????????????????package
\usepackage{color}
\usepackage[normalem]{ulem}
 \usepackage{indentfirst}
 \usepackage{times}
\usepackage{epsfig}
\usepackage{stmaryrd}
\usepackage{diagbox}
\usepackage{comment}
 \usepackage{slashbox}

\usepackage{hyperref}
\usepackage{amsmath,amssymb,amsfonts,amsthm}
\usepackage{algorithm}
\usepackage{algorithmic}

\newcounter{ToDo}
\newcounter{gaocomm}
\newcounter{Note}
\definecolor{blue-violet}{rgb}{0.54, 0.17, 0.89}
\definecolor{mygreen}{rgb}{0.0, 0.5, 0.0}
\definecolor{awesome}{rgb}{1.0, 0.13, 0.32}
\definecolor{bostonuniversityred}{rgb}{1.0, 0.0, 0.0}
\begin{document}
\title{\textbf{Vectorial Dimension Reduction for Tensors Based on Bayesian Inference}}
%\markboth{IEEE Transactions on Neural Networks and Learning Systems, Vol.~XX, No.~X, XXX~2016}%
%{F. Ju \MakeLowercase{\textit{et al.}}: Tensor Bayesian ...}

\author{Fujiao Ju,
        Yanfeng Sun,
        Junbin Gao,
        Yongli Hu and Baocai Yin
\thanks{Fujiao Ju, Yanfeng Sun and Yongli Hu are with Beijing Advanced Innovation Center for Future Internet Technology, Beijing Key Laboratory of Multimedia and Intelligent Software Technology, Beijing University of Technology, Beijing 100124, P. R. China, e-mail:jufujiao2013@emails.bjut.edu.cn, \{yfsun, huyongli\}@bjut.edu.cn, Corresponding author is Yanfeng Sun. }
\thanks{Junbin Gao is with the Discipline of Business Analytics, The University of Sydney Business School, The University of Sydney, NSW 2006, Australia. \protect e-mail: junbin.gao@sydney.edu.au}
\thanks{Baocai Yin is with the Faculty of Electronic Information and Electrical Engineering, Dalian University of Technology, e-mail:ybc@dlut.edu.cn}}
%\thanks{Manuscript received January 1, 2015; revised December XX, 2015.}

\maketitle

\begin{abstract}
Dimensionality reduction for high-order tensors is a challenging problem. In conventional approaches, higher order tensors are ``vectorized'' via Tucker decomposition to obtain lower order tensors. This will  destroy the inherent high-order structures or resulting in undesired tensors, respectively. This paper introduces a probabilistic vectorial dimensionality reduction model for tensorial data. The model represents a tensor by employing a linear combination of same order basis tensors, thus it offers a mechanism to directly reduce a tensor to a vector. Under this expression, the projection base of the model is based on the tensor CandeComp/PARAFAC (CP)  decomposition and the number of free parameters in the model only grows linearly with the number of modes rather than exponentially. A Bayesian inference has been established via the variational EM approach. A criterion to set the parameters (factor number of CP decomposition and the number of extracted features) is empirically given. The model outperforms several existing PCA-based methods and CP decomposition on several publicly available databases in terms of classification and clustering accuracy.
\end{abstract}


% \fi
\IEEEpeerreviewmaketitle

\section{Introduction}\label{Sec:1}
High-dimensional and multiple array data are widely acquired in modern computer vision research \cite{JuSunGaoHu2016, KoniuszCherian2015, SoltaniKilmerHansen2015}. This high- and multi-dimensional data normally lie close to a manifold of much lower dimension \cite{TorkiElgammalLee2010, ZhangHuangWang2010}. It has been a challenging problem to find low-dimensional embedding for high-dimensional observed data in machine learning research, although great progresses have been made for dimensionality reduction in the last couple of decades \cite{CelikLogsdonAase2014,JenattonObozinskiBach2010,  KongWangTeohWangVenkateswarlu2005,  LuPlataniotisVenetsanopoulos2008,LuLaiFanCuiZhu2015,WongLaiXuWenHo2015}.

Principal component analysis (PCA) \cite{Bishop2006, Jolliffe2002,KunchevaFaithfull2014} is one of popular dimensionality reduction methods widely used in image analysis \cite{HoyerHyvarinen2000,KeSukthankar2004}, pattern recognition \cite{ChenDaiPanHuang2015,HeYanHuNiyogiZhang2005, LuPlataniotisVenetsanopoulosLi2006} and machine learning \cite{KriegelKroegerSchubert2008} for data analysis. As a well-known dimensionality reduction method for vectorial data, PCA represents vectorial data by using a linear combination of the basis vectors (principal directions), which have same dimensionality as vectorial data. Due to the mutual orthogonality of basis vectors, the weight coefficient can be taken as the dimension-reduced representation of vectorial data. However, contemporary data emerging from science and technology are in new types with more structures. For example, an image or video should be regarded as 2D or 3D data in order to preserve pixel spatial information. While conducting dimensionality reduction for images or videos by the classical PCA, a typical work around is to vectorize data. Vectorizing 2D/3D data not only results in very high-dimensional data, causing the curse of dimensionality \cite{XieYanKwokHuang2008}, but also ignores the important spatial relationship between features within image or video. Instead of using vectorization, some 2D or tensor dimensionality reduction method have been introduced. For 2D data, several 2DPCA algorithms have been proposed \cite{JuSunGaoHu2016,JuSunGaoHuYin2015,YangZhangFrangiYang2004}. All these methods reduce respectively the data dimension in the row or column direction and the dimension-reduced representation of 2D data is still in 2D.
For high order tensors, the typical ways of dimensionality reduction are  the CP and Tucker decomposition
 \cite{KoldaBader2009}. %The multidimensional data can be seen as tensors instead of matrices or vectors.
The CP decomposition can be interpreted as a sum of $R$ rank-one tensors ($R$ represents CP's rank) \cite{ZhaoZhouZhangCichockiAmari2016}. Tucker decomposition can be seen as $N$-order singular value decomposition (SVD) \cite{AlexVasilescuTerzopoulos2003}. These algorithms reduce the dimension from each tensor mode. When applying CP and Tucker on a tensor, which is stacked from $N$ samples, the loading matrix of the last mode  can be taken as extracted
features of these samples with reduced dimensions. Thus, for CP decomposition, the feature number is equal to CP's rank, which is a strong restriction. For Tucker decomposition, it produces a dimension-reduced representation being still a tensor with same order as original data.

 %But they produce the dimension-reduced representation being still tensors in the same order as original data. Sometimes this is not appropriate for sequence data analysis.

The majority of aforementioned models falls in the  algebraic  paradigm. Algebraic model don't always have flexibility of providing confidence information of the model when dealing with noisy data. To combat this drawback in the case of vectorial data, Tipping and Bishop \cite{TippingBishop1999} proposed a probabilistic PCA model, called PPCA. Under the probabilistic learning framework, the model parameters in PPCA can be easily solved by EM approach. %Zhao \emph{et al}. \cite{ZhaoYuKwok2012} proposed a probabilistic principle component analysis model on 2D data based on Gaussian noise assumption, called 2DPPCA, which takes a step forward for PPCA model in 2D cases.
Being concerned with a probabilistic PCA model for Laplace noise, Gao \cite{Gao2007} introduced a robust probabilistic PCA model based L1-norm, called L1-PPCA. Ju \emph{et al} \cite{JuSunGaoHuYin2015} proposed a dimensionality reduction  algorithm via L1 norm-based 2D probabilistic PCA. The authors expressed the Laplacian distribution as a superposition of an infinite number of Gaussian densities, with precision controlled by another latent variable $\beta$ which can be seen
as an indicator for outliers.

Inspired by  probabilistic latent factor models \cite{MnihSalakhutdinov2007,SalakhutdinovMnih2008}, some tensor Bayesiant factorization have been proposed \cite{ErmisTaylanCemgil2014,  HayashiTakenouchiShibataKamiyaKatoKuniedaYamadaIkeda2010,RaiWangGuoChenDunsonCarin2014,YangDunson2015}. Xiong \emph{et al}. \cite{XiongChenHuangSchneiderCarbonell2010} applied a full Bayesian treatment to derive an almost parameter-free probabilistic tensor factorization algorithm called as Bayesian Probabilistic Tensor Factorization
(BPTF). Zhao \emph{et al.} \cite{ZhaoZhangCichocki2015} formulate CP decomposition  using a hierarchical probabilistic model and employ a fully Bayesian treatment by incorporating a sparsity-inducing prior over multiple latent factors and the appropriate hyperpriors over all the hyperparameters, resulting in automatic rank determination. And then in \cite{ZhaoZhangCichocki2015}, they presented a class of probabilistic generative Tucker models for tensor decomposition and completion with
structural sparsity over multilinear latent space.
%Morup and Hansen \cite{MorupHansen2009} demonstrated how a Bayesian framework for model selection based on automatic relevance determination (ARD) can be adapted to the Tucker and CandeComp/PARAFAC (CP) models.

While applying CP-based approaches in feature extraction, the number of features is implicitly determined by CP's rank, i.e., the number of rank-1 tensorial bases. This is a strong restriction. If the rank is too small, the representation power of the CP decomposition is limited; if the rank is too large, the model will overfit the training data. For Tucker-based approaches, they are usually used to clustering \cite{SunGaoHongMishraYin2015}. In addition, the produced dimension-reduced representations are still tensors with the same order as original data. The tensor feature will cannot be used easily for image analysis. %\GaoR{I will rewrite the last several statements here.}

To address these issues, in this paper, we propose   a new tensor dimensionality reduction model based on Bayesian theory. This model represents a tensor as a linear combination of some tensor bases in the same order with the coefficient vector being taken as the dimension-reduced representation. It is more similar to generalizing the vectorial PCA to the tensorial case, where the noise model has been considered as the structured matrix-variate Gaussian. But reducing dimension to the vectorial type has one problem: the number of parameters to be estimated in the tensor bases will increase significantly.  The similar phenomenon has been observed in neural network research. For example, in \cite{NovikovPodoprikhinOsokinVetrov2015}, the authors constrained the parameters in the neural network with a tensor train (TT) form. %The TT-layer consists in storing the weights matrtix of the fully-connected layer in the TT-format, allowing to use hundreds of thousands (or even millions) of hidden units while having moderate number of parameters.
Nguyen \emph{et al} \cite{NguyenTranPhungVenkatesh2014} introduced Tensor-variate Restricted Boltzmann Machine (TvRBM) which generalizes RBM to caputure the multiplicative interaction between data modes and the latent variables. To avoid the number of mapping free parameters from growing too fast, the authors employed $(N+1)$-way factoring to construct the multiplicative interactions between visible modes and hidden units. Thus TvRBMs are highly compact in the sense that the number of free parameters grows only linearly with the number of modes. %In paper \cite{QiGaoSunHuLi2015}, the authors specified a multiplicative interaction between visible unit and hidden unit by taking outer-product of matrices.
Motivated by these idea, we constrain the basis tensor with CP structure in our model and then learn a set of rank-1 bases for a group of tensor data.
%use some basic $N$th-order tensors to construct a $(N+1)$th-order tensor, and start from a similar decomposition to\cite{NguyenTranPhungVenkatesh2014},  e.g. CP decomposition.

The proposed tensor dimensionality reduction model has the following advantages: Firstly, we can use the proposed model to extract features on tensor data directly and the obtained feature is of vector instead of high-order tensor data. Secondly, {by constraining the basis tensor with CP structure, the number of parameters to be estimated grows linearly with the number of modes rather than exponentially. Thirdly, in the proposed model, the feature number is determined by dimension of coefficient vector, while the flexibility of subspace basis is determined by the CP rank. This implies that we can obtain very different features even the number of features is fixed while varying the CP s rank. Therefore, the proposed model might provide a flexible feature extraction framework compared with CP decomposition. A criterion to set the feature number and the CP rank is given in our experiments and the performance of the proposed model is assessed through the classification and clustering experiments on real world databases.

The rest of this paper is organized as follows. Section \ref{Sec:2} introduces some basic algebraic notations and the concepts of tensorial decomposition. In Section \ref{Sec:3}, we introduce the second-order tensor Bayesian vectorial dimensionality reduction model and give the derivation of the variational approximation technology for solving the proposed model. In Section \ref{Sec:4}, we extend the second-order tensor Bayesian vectorial dimensionality reduction model to high-order cases. In Section \ref{Sec:5}, experimental results are presented to evaluate the performance of the proposed model. Finally, conclusions and future works are summarized in Section \ref{Sec:6}.

%-------------------------------------------------------------------------
\section{Preliminaries}\label{Sec:2}
A tensor is a multidimensional array \cite{KoldaBader2009}. More formally, an $N$-order tensor is an element of the tensor product of $N$ vector spaces, each of which has its own coordinate system. It is higher-order generalization of scalar (zeroth-order tensor), vector (first-order tensor), and matrix (second-order tensor). In this paper, lowercase italic letters ($a$, $b$, $\cdots$) denote scalars, boldface lowercase letters ($\mathbf a$, $\mathbf b$, $\cdots$) denote vectors, boldface uppercase letters ($\mathbf A$, $\mathbf B$, $\cdots$) denote matrices, and boldface Euler script letters ($\mathcal A$, $\mathcal B$, $\cdots$) denote tensors. Generally an $N$-order tensor is given by $\mathcal{A} = (a_{d_1d_2\cdots d_N})$ with $N$ indices satisfying $1\leq d_k \leq D_k$ ($k=1,2,...,N$), denoted by $\mathcal{A}\in\mathbb{R}^{D_1\times\cdots\times D_N}$.  $(D_1, ..., D_N)$ is called the dimension of $\mathcal{A}$.
%In the paper, we denote vectors by boldface lowercase letters, e.g., $\mathbf x$. Matrices are denoted by boldface capital letters, e.g., $\mathbf X$. The $j$th column of $\mathbf X$ is denoted by $\mathbf x_{:,j}$, and the row of a matrix $\mathbf X$ is denoted by $\mathbf x_{i,:}$. Alternatively, the $j$th column of a matrix, $\mathbf x_{:,j}$, may be denoted more compactly as $\mathbf x_j$. $N$th-order tensors (multi-way arrays) are denoted by boldface Euler script letters, e.g.,$\mathcal X$. Slices are two-dimensional sections of a tensor, defined by fixing all but two indices. The horizontal, lateral, and frontal slides of a third-order tensor $\mathcal X$, denoted by $\mathbf X_{i,:,:}$, $\mathbf X_{:,j,:}$, and $\mathbf X_{:,:,k}$, respectively. Alternatively, the $k$th frontal slice of a third-order tensor, $\mathbf X_{:,:,k}$ may be denoted more compactly as $\mathbf X_k$.

The inner product of two $N$-order tensors $\mathcal A,\mathcal B\in\mathbb R^{D_1\times D_2\times\cdots\times D_N}$ is the sum of the products of their corresponding entries, i.e.,
\[
\langle\mathcal A,\mathcal B\rangle = \sum_{d_1=1}^{D_1}\sum_{d_2=1}^{D_2}\cdots\sum_{d_N=1}^{D_N}a_{d_1d_2\cdots d_N}b_{d_1d_2\cdots d_N}.
\]

The CP decomposition of an $N$-order tensor means that it can be factorized into a sum of component rank-one tensors.
\begin{align}
\mathcal A&\approx \sum_{r=1}^R\lambda_r\mathbf a_r^{(1)}\circ\mathbf a_r^{(2)}\circ\cdots\circ\mathbf a_r^{(N)}\label{CP};\\
\nonumber&= \llbracket\boldsymbol{\lambda};\mathbf A^{(1)},\mathbf A^{(2)},\ldots,\mathbf A^{(N)}\rrbracket,
\end{align}
where $\circ$ means the outer product of vectors. $R$ is a given positive integer, $\boldsymbol{\lambda}\in\mathbb R^{R}$ and $\mathbf A^{(n)}=[\mathbf a^{(n)}_1,\mathbf a^{(n)}_2,...,\mathbf a^{(n)}_R]\in\mathbb R^{D_n\times R}$ for $n=1,2,\ldots,N$ are factor matrices. Each component on the right hand side of \eqref{CP} is a rank-one tensor in the same order as $\mathcal{A}$.
%The factor matrices refer to the combination of the vectors from the rank-one components, i.e., $\mathbf A^{(n)}=[\mathbf a^{(n)}_1,\mathbf a^{(n)}_2,...,\mathbf a^{(n)}_R]$.
Elementwise, \eqref{CP} is written as
\[
a_{d_1,d_2,\ldots,d_N}\approx\sum_{r=1}^R\lambda_ra_{d_1r}^{(1)}a_{d_2r}^{(2)}\cdots a_{d_nr}^{(N)},
\]
for $1\leq d_n \leq D_n$. In this case, the mode-$n$ matricized version is given by
\[
\mathbf A_{(n)}\approx\mathbf A^{(n)}\boldsymbol{\Lambda}(\mathbf A^{(N)}\circledast\cdots\circledast\mathbf A^{(n+1)}\circledast\mathbf A^{(n-1)}\circledast\cdots\mathbf A^{(1)})^T,
\]
where $\boldsymbol{\Lambda}=\text{Diag}(\boldsymbol{\lambda})$, ``Diag" represents generating the diagonal matrix from the vector $\boldsymbol{\lambda}$ and $\circledast$ is Khatri-Rao product, which presents matching columnwise ¡± Kronecker product
  \cite{KoldaBader2009}.

Without loss of generality, we can assume that $\boldsymbol{\Lambda}=\mathbf I$ to be the identity matrix by absorbing all the elements of $\boldsymbol{\lambda}$ into those $\mathbf A^{(n)}$. Particularly, given a third-order tensor $\mathcal X\in \mathbb R^{D_1\times D_2\times D_3}$, we can write it as
\[
\mathcal X\approx \sum_{r=1}^{R}\mathbf a_r\circ \mathbf b_r\circ \mathbf c_r = \llbracket\mathbf A,\mathbf B,\mathbf C\rrbracket,
\]
where $\mathbf a_r\in\mathbb R^{D_1}$, $\mathbf b_r\in\mathbb R^{D_2}$ and $\mathbf c_r\in\mathbb R^{D_3}$ for $r=1,\ldots,R$, and $\mathbf{A} = [\mathbf{a}_1, ..., \mathbf{a}_R]$, $\mathbf{B} = [\mathbf{b}_1, ..., \mathbf{b}_R]$ and $\mathbf{C} = [\mathbf{c}_1, ..., \mathbf{c}_R]$. The three-way model  can also be written as, in terms of the frontal slices of $\mathcal X$,
\[
\mathbf X_{d_3}\approx\mathbf A\mathbf D^{(d_3)}\mathbf B^T,
\]
where $\mathbf D^{(d_3)}\equiv \text{Diag}(\mathbf c_{d_3,:})$ and $\mathbf c_{d_3,:}$ represents the $d_3$-th row of matrix $\mathbf C$ for $1\leq d_3\leq D_3$.

According to the tensor theory, the CP decomposition of a  tensor $\mathcal X$ is not uniquely identifiable due to the elementary indeterminacy of scaling and permutation. The permutation indeterminacy refers to the fact that the rank-one component tensors can be ordered arbitrarily, i.e.,
\[
\mathcal X = \llbracket \mathbf A,\mathbf B,\mathbf C \rrbracket= \llbracket \mathbf {A\Pi},\mathbf {B\Pi},\mathbf{C\Pi}\rrbracket,
  \]
for any $R\times R$ permutation matrix $\mathbf{\Pi}$. The scaling indeterminacy refers to the fact that we can scale the individual vectors, i.e.,
\[
\mathcal X = \sum_{r=1}^R(\alpha_r\mathbf a_r)\circ(\beta_r\mathbf b_r)\circ(\gamma_r\mathbf c_r),
\]
as long as $\alpha_r\beta_r\gamma_r=1$ for $r=1,\ldots,R$.

%-------------------------------------------------------------------------
\section{Second-Order Tensor Bayesian Vectorial Dimension Reduction Model}\label{Sec:3}
The goal of this paper is to consider tensor Bayesian vectorial dimensionality reduction model. In order to explain our model clearly, we first consider the second-order tensor (matrix) dimensionality reduction model, then extend the model to the high-order cases.

\subsection{The Proposed Model}
Given a sample set $\{\mathbf {Y}_i|\mathbf {Y}_i\in \mathbb R^{D_1\times D_2},i =1,\ldots,M\}$, which contains $M$ independently and identically distributed samples in $\mathbb{R}^{D_1\times D_2}$. These samples can form a third-order tensor $\mathcal Y\in \mathbb R^{D_1\times D_2\times M}$ with every frontal slice of $\mathcal Y$ being a sample $\mathbf {Y}_i$. The proposed model assumes
that each $\mathbf {Y}_i$ can be additively decomposed as a linear latent variable model and noise, that is

\begin{equation}\label{model}
\mathbf {Y}_i=\mathcal{W}\times_3 \mathbf {h}_i^T+\mathbf {E}_i,\quad i=1,\ldots,M,
\end{equation}
where $\mathcal{W}\in \mathbb{R}^{D_1\times D_2\times K}$, $\times_3$ denotes the product of a tensor and a vector \cite{KoldaBader2009}, $\mathbf h =\{\mathbf {h}_i\}_{i=1}^M$ with $\mathbf {h}_i\in \mathbb{R}^K$ and $K$ represents the reduced-dimension. In other words, the model (\ref{model}) can also be presented as
\[
\mathbf Y_i = \sum_{k=1}^K h^{(i)}_k\mathbf {W}_k+\mathbf {E}_i,
\]
where $h^{(i)}_k$ represents the $k$th element of $\mathbf {h}_i$ and $\mathbf {W}_k$ is the $k$th frontal slice of tensor $\mathcal W$. In this case, we have rewritten each sample $\mathbf {Y}_i$ as a linear combination of projection bases $\mathbf {W}_k(k=1,\ldots,K)$. The projection base $\mathbf {W}_k$ has the same size as the sample $\mathbf {Y}_i$. Thus we can obtain the vectorial dimension reduction for a 2D data.

We assume that the noise $\mathbf{E}_i$ satisfies a matrix-variate Gaussian distribution $\mathcal{N}(0,\sigma \mathbf I,\sigma \mathbf I)$ \cite{XuYanQi2012}. This means that each component $e^{(i)}_{d_1,d_2}$ of $\mathbf {E}_i$ follows normal distribution $\mathcal{N}(0,\sigma^2)$. To develop a generative Bayesian model, we further impose a prior on the latent variable,
\[
p(\mathbf {h}_i)=\mathcal{N}(\mathbf {h}_i|0,\mathbf I_K)=(\frac{1}{2\pi})^{K/2}\exp\{-\frac{1}{2}\mathbf {h}_i^T\mathbf {h}_i\}.
\]
For simplicity, we impose a Gamma prior on $\rho = \frac{1}{\sigma^2}$ instead of directly on $\sigma$. The prior is given by
\[
p_{\sigma}(\rho)=\Gamma(\rho|a,b)=\frac{b^a}{\Gamma(a)}\rho^{a-1}\exp\{-b\rho\}.
\]

For the proposed model, in order to introduce the variational learning method, we assume $\mathbf h$ and $\rho$ are the model hidden variables and $\mathcal W$ is a parameter. For the given observation $\mathcal Y$, maximizing the likelihood $p(\mathcal Y|\mathcal W)$ as a function of $\mathcal W$ is equivalent to
maximizing the log likelihood
\begin{align*}
&\mathcal{L}(\mathcal W)=\log p(\mathcal Y|\mathcal W) = \log \int p(\mathcal Y, {\mathbf h}, \rho|\mathcal W)d\mathbf h d\rho,
\end{align*}
where the joint distribution is given by
\begin{align*}
& p(\mathcal Y, {\mathbf h}, \rho|\mathcal W) \\
=& \prod_{i=1}^M\mathcal{N}(\mathbf Y_i-\mathcal{W}\times_3 \mathbf h_i|0, \sigma \mathbf I,\sigma\mathbf I)\mathcal{N}(\mathbf h_i|0, \mathbf I_K)p_{\sigma}(\rho).
\end{align*}

%-------------------------------------------------------------------------
\subsection{Variational EM Algorithm}
For the third-order tensor $\mathcal Y$ formed from all the given samples $\{\mathbf {Y}_i|\mathbf {Y}_i\in \mathbb {R}^{D_1\times D_2},i =1,\ldots,M\}$ along the third mode, the learning task is to learn the model parameter $\mathcal W$ such that the log likelihood function is maximized. Using any distribution $Q(\mathbf {h},\rho)$, called variational distribution over the hidden variables, we can obtain a lower bound on $\mathcal L(\mathcal W)$:
\begin{align}
\nonumber &\mathcal L(\mathcal W) = \log p(\mathcal Y|\mathcal W) \\
\nonumber\geq &\int_\mathbf {h} \int_{\rho}Q(\mathbf {h}, \rho)\log\frac{p(\mathcal Y,\mathbf {h},\rho|\mathcal{W})}{Q(\mathbf {h},\rho)}d\mathbf {h} d\rho\\
 =& E_{\mathbf {h},\rho}[\log p(\mathcal Y|\mathbf {h},\rho, \mathcal{W})]+E_{\mathbf h}[\log\frac{p(\mathbf h)}{Q(\mathbf h)}]+E_{\rho}[\log\frac{p(\rho)}{Q(\rho)}] \label{Exp}\\
\nonumber\triangleq &\mathcal F(Q(\mathbf h), Q(\rho),\mathcal W).
\end{align}
The above inequality is based on Jensen's inequality, referring to \cite{Bishop2006} for more details. The second equation is based on the assumption that $Q(\mathbf h,\rho)$ is separable, e.g., $Q(\mathbf h,\rho)= Q(\mathbf h)Q(\rho)$.

The purpose of the variational EM \cite{Bishop2006} is to maximize $\mathcal F(Q(\mathbf h), Q(\rho),\mathcal W)$ with respect to $Q(\mathbf h), Q(\rho)$ and $\mathcal W$. A complete derivation is given in APPENDIX.

\subsubsection{\textbf{Variational E-step}} In E-step, we update Q-distributions of all the hidden variables with the current fixed parameter values for $\mathcal W$.
\\[2mm]
\noindent(1) Update the posterior of $\mathbf{h}_i$:

Given $\mathbf{Y}_i$, we can verify that the best approximated Q-distribution of $\mathbf{h}_i$ is the normal distribution $\mathcal{N}(\mathbf {u}_i,\mathbf{\Sigma})$ with appropriate mean $\mathbf{u}_i$ and covariance $\mathbf{\Sigma}$. To see this, first note that,
in (\ref{Exp}), the last expectation term is constant with respect to $\mathbf {h}_i$, hence we are only concerned with the computation of the first two expectations. Thus, the lower bound attains its maximum at the normal distribution with variational parameters $\mathbf{u}_i$ and $\mathbf \Sigma$ are given, respectively, as follows,
\begin{align}
\mathbf{u}_i = (\mathbf{\Sigma_{\mathcal{W}}}+1/\overline{\rho}\mathbf I_K)^{-1}\mathbf a_i\label{meanh}
\end{align}
and
\begin{align}
\mathbf \Sigma = (\mathbf I_K+\overline{\rho}\mathbf{\Sigma_{\mathcal{W}}})^{-1},\label{variance}
\end{align}
where $\mathbf \Sigma_{\mathcal W}$ is a $K\times K$ symmetric matrix with the $pq$-element $\text{tr}(\mathbf W_p^T\mathbf W_q)$, $p,q=1,\ldots,K$, $\mathbf{a}_i$ is a $K\times 1$ vector with the element $\text{tr}(\mathbf W_k^T\mathbf Y_i)$, $k=1,\ldots,K$, and $\overline{\rho}$ is the mean of $\rho$ with respect to the approximate posterior $Q(\rho)$.
\\[2mm]
\noindent(2) Update the approximated  posterior of $\rho$:

Under the framework of variational inference, the best distribution $Q^*(\rho)$ can be calculated as
\begin{align*}
\log Q^*(\rho)&=E_{\mathbf h}[\log p(\mathcal Y,\mathbf h,\rho|\mathcal W)]\\
&=\frac{D_1D_2M}{2}\log\rho-\frac{\rho}{2}\psi+(a-1)\log\rho-b\rho,
\end{align*}
where
\[
\psi=M\text{tr}(\mathbf \Sigma_{\mathcal W}\mathbf\Sigma)+\sum_{i=1}^M\|\mathbf Y_i-\mathcal W\times_3\mathbf u_i^T\|_F^2.
\]
It demonstrates that the log of the optimal solution for the latent variable $\rho$ is obtained by simply considering the log of the joint distribution over all hidden and visible variables and then taking the expectation with respect to all the other latent variables. From the above equation, we can get
\[
Q^*(\rho)\propto \rho^{a-1}+\frac{D_1D_2M}{2}\exp\{-b\rho-\frac{\psi}{2}\rho\}.
\]
Hence the best $Q^*(\rho)$ is still a Gamma distribution $\Gamma(\rho|\overline{a},\overline{b})$ but with the updated parameters:
\begin{align}
\overline{a}=a+\frac{D_1D_2M}{2}\quad \text{and}\quad \overline{b}=b+\frac{1}{2}\psi.\label{updateAB}
\end{align}

\subsubsection{\textbf{Variational M-step}}
In M-step, with the variational distributions fixed at $Q$'s, we need update the parameter $\mathcal W$ to maximize $\mathcal F(Q,\mathcal W)$. A major problem with the projection tensor in (\ref{model}) is the excessively large number of free parameters. If $\mathcal W$ is a $(N+1)$-order projection tensor with $K\prod_nD_n$ elements, it quickly reaches billions when the mode dimensionalities $K$, $D_{1:N}$ and $N$ are moderate. This makes parameter learning of base tensor extremely difficult. So we can employ $(N+1)$-order factoring \cite{NguyenTranPhungVenkatesh2014} to construct the multiplicative interactions in tensor $\mathcal W$. With $R$ factors, we restrict ourselves to the tensorial parameterization of $\mathcal W$ in the CP decomposition as follows:
\begin{align}
\nonumber\mathcal{W}&=\llbracket\boldsymbol{\lambda};\mathbf W^{(1)},\mathbf W^{(2)},\mathbf W^{(h)}\rrbracket\\
&= \sum_{r=1}^{R}\lambda_r\mathbf w^{(1)}_{:,r}\circ\mathbf w^{(2)}_{:,r}\circ\mathbf w^{(h)}_{:,r},\label{W-definition}
\end{align}
where $\boldsymbol{\lambda}$ is the scaling vector, the factor matrix $\mathbf W^{(n)}\in\mathbb{R}^{D_n\times R}$ $(n=1,2)$ and $\mathbf W^{(h)}\in\mathbb{R}^{K\times R}$. For simplicity, we fix $\boldsymbol{\lambda}=1$, so we obtain:
\[
w_{d_1d_2k}=\sum_{r=1}^Rw^{(1)}_{d_1r}w^{(2)}_{d_2r}w^{(h)}_{kr}.
\]
Now $\mathbf \Sigma_{\mathcal W}$ becomes
\[
 \mathbf\Sigma_{\mathcal W}=\mathbf W^{(h)}[(\mathbf W^{(1)T}\mathbf W^{(1)})\odot(\mathbf W^{(2)T}\mathbf W^{(2)})] \mathbf W^{(h)T},
\]
where $\odot$ means the elementwise product.
The details are shown in Supplementary Material.
To solve for such parameterized $\mathcal W$, we gather all the terms related to $\mathcal W$ in (\ref{Exp}) and get
\begin{align}
\nonumber{\mathcal F}&\propto E_{\mathbf h,\rho}[\log p(\mathcal Y|\mathbf h,\rho,\mathcal W)]\\
\nonumber=&\sum_{i=1}^M\|\mathbf Y_i-\mathcal W\times_3\mathbf u_i^T\|_F^2+M\text{tr}(\mathbf \Sigma_{\mathcal W}\mathbf \Sigma)+\text{Const}\\
\nonumber=&\sum_{i=1}^M\|\mathbf Y_i-\mathbf W^{(1)}\text{Diag}(\mathbf u_i^T\mathbf W^{(h)})\mathbf W^{(2)T}\|_F^2\\
\nonumber&+M\text{tr}([(\mathbf W^{(1)T}\mathbf W^{(1)})\odot(\mathbf W^{(2)T}\mathbf W^{(2)})] \mathbf W^{(h)T} \mathbf\Sigma\mathbf W^{(h)})\\
\triangleq &\mathcal F^{'}\label{ApproxF1}
\end{align}

As we know, for any matrices $\mathbf A$, $\mathbf B$ and $\mathbf C$, we have
 \[
 \frac{\partial\text{tr}[(\mathbf A^T\mathbf A)\odot(\mathbf B^T\mathbf B)\mathbf C]}{\partial\mathbf A}= \mathbf A((\mathbf B^T\mathbf B)\odot\mathbf C+(\mathbf B^T\mathbf B)\odot\mathbf C^T).
 \]
Thus, maximizing the lower bound $\mathcal F^{'}$ with respect to factor loadings $\mathbf W^{(1)}$, $\mathbf W^{(2)}$ and $\mathbf W^{(h)}$, we can obtain
\begin{align}
\nonumber&\mathbf W^{(1)}=\\
\nonumber&[\sum_{i=1}^N\mathbf Y_i\mathbf W^{(2)}\text{Diag}(\mathbf u_i^T\mathbf W^{(h)})]\\
\nonumber&[(\mathbf W^{(2)T}\mathbf W^{(2)})\odot(M\mathbf W^{(h)T} \mathbf\Sigma\mathbf W^{(h)}+\mathbf W^{(h)T}\mathbf U\mathbf U^T\mathbf W^{(h)})]^{-1}. \\
\label{W1}
\end{align}
\begin{align}
\nonumber&\mathbf W^{(2)}=\\
\nonumber&[\sum_{i=1}^M\mathbf Y_i^T\mathbf W^{(1)}\text{Diag}(\mathbf u_i^T\mathbf W^{(h)})]\\
\nonumber&[(\mathbf W^{(1)T}\mathbf W^{(1)})\odot(\mathbf W^{(h)T} \mathbf\Sigma\mathbf W^{(h)}+\mathbf W^{(h)T}\mathbf U\mathbf U^T\mathbf W^{(h)})]^{-1}\\
\label{W2}
\end{align}
and
\begin{align}
\nonumber\mathbf W^{(h)}=&[\mathbf U\mathbf U^T+M\mathbf\Sigma]^{-1}[\sum_i\mathbf u_i[\text{diag}(\mathbf W^{(2)T}\mathbf Y_i^T\mathbf W^{(1)})]^T]\\
&[(\mathbf W^{(1)T}\mathbf W^{(1)})\odot(\mathbf W^{(2)T}\mathbf W^{(2)})]^{-1}\label{Wh}
\end{align}
where ``diag'' generates a vector using diagonal elements of the matrix.

\subsubsection{\textbf{The Overall Algorithm}}
In variational E-step, we update Q-distributions of all the hidden variables with the current fixed parameter value $\mathcal W$. In variational M-step, we fix all the distributions over the hidden variables and update the parameter $\mathbf W^{(1)}$, $\mathbf W^{(2)}$ and $\mathbf W^{(h)}$. The two steps are alternatively continued until a termination condition is satisfied.

We define the restruction errors between original image set and the restructured images set as
\begin{align}
e(t)= 1-\frac{\|\mathcal Y - \mathcal W_{t}\times_3\mathbf U_{t}^T\|_F}{\|\mathcal Y\|_F},\label{define_e}
\end{align}
where $\mathbf U_t$ is made up of all $\mathbf u_i$ in $t$-th step.
To terminate the iteration, up to a given maximum iterative number $T$, we set stopping condition satisfies $|e(t)-e(t+1)|<\epsilon$, where $\epsilon$ is a given value.

The above variational EM algorithm is summarized in Algorithm 1.
\begin{algorithm}
\renewcommand{\algorithmicrequire}{\textbf{Initialize:}}
\renewcommand\algorithmicensure {\textbf{Variational E-step:} }
\renewcommand\algorithmicensure {\textbf{Variational M-step:} }
\caption{Second-Order Tensor Bayesian Vectorial Dimension Reduction (TBV-DR) Algorithm}
\begin{algorithmic}[1]
  \REQUIRE Training set $\{\mathbf Y_i, i=1,\ldots,M\}$; Initialize factor matrices $\mathbf W^{(1)}$, $\mathbf W^{(2)}$ and $\mathbf W^{(h)}$, Gamma parameters $a$, $b$ and $\epsilon$.
 \FOR {$t=1$ to $T$}
    %\ENSURE
    \STATE \textbf{Variational E-step:}
    \STATE Maximize the lower bound $\mathcal F$ with respect to $Q(\mathbf h)$: Iterate the variational parameters $\mathbf u_i$ and $\mathbf \Sigma$ in $Q(\mathbf h_i)$ based on (\ref{meanh}) and (\ref{variance}) for all $i=1,...,M$.\\
    \STATE Maximize the lower bound $\mathcal F$ with respect to $\rho$: Iterate the variational parameters $a$ and $b$ based on (\ref{updateAB})\\
     \STATE \textbf{Variational M-step:}
    \STATE Maximize the lower bound $\mathcal F^{'}$ with respect to $\mathbf W^{(1)}$, $\mathbf W^{(2)}$ and $\mathbf W^{(h)}$: update $\mathbf W^{(1)}$ based on (\ref{W1}), $\mathbf W^{(2)}$ based on (\ref{W2}) and $\mathbf W^{(h)}$ based on (\ref{Wh}).
    \STATE\textbf{calculate }$e(t)$.
   \STATE\textbf{if } $|e(t)-e(t+1)|<\epsilon$,   break;   \textbf{end}

 \ENDFOR
\end{algorithmic}
\end{algorithm}
%-------------------------------------------------------------------------
\section{High-Order Tensor Bayesian Vectorial Dimension Reduction Model}\label{Sec:4}
\begin{comment}
\begin{table*}[htb]
\renewcommand{\arraystretch}{1.5}
\footnotesize
\centering
\begin{tabular}{|c|c|c|c|c|c|}
  \hline
Methods &Model &Storage Space\\
\hline
  GLRAM &$\mathcal Y =\mathcal Z\times_1 \mathbf L\times_2\mathbf R$ &$(D_1D_2...D_{(N-1)})\times r+D_N\times c+r\times c\times M$ \\
  \hline
TUCKER & $\mathcal Y=\mathcal G\times_1 \mathbf U^{(1)}...\times_N \mathbf U^{(N)}\times_{(N+1)}\mathbf U^{(m)}$& $\sum_{n=1}^ND_nR_n+MP+P\prod_{n=1}^N R_n$ \\
  \hline
TBV-DR& $\mathcal Y =\mathcal W\times_{(N+1)}\mathbf H^T$ &$(K+\sum_{n=1}^N D_n)F+MK$\\
\hline
\end{tabular}
    \caption{Storage space of different algorithms.}\label{storage}
\end{table*}
\end{comment}
In this section, we extend the second order TBV-DR algorithm to the case of high-order tensor.
\subsection{The high-order TBV-DR model}
For a given set of $N$-order tensor samples $\{\mathcal Y_i|\mathcal Y_i\in\mathbb R^{D_1\times D_2\times ...\times D_N}, i=1\ldots,M\}$, stack them into a $(N+1)$-order tensor $\mathcal Y\in \mathbb R^{D_1\times D_2\times\ldots\times D_N \times M}$. Then the model in (\ref{model}) becomes:
\[
\mathcal Y_i = \mathcal W\times_{(N+1)}\mathbf h_i^T+\mathcal E_i,
\]
where $\mathbf h_i\in \mathbb R^K$ and $\mathcal W\in \mathbb R^{D_1\times D_2\times\ldots\times D_N\times K}$ is  an $(N+1)$-order tensor. Now we define $\mathcal W_k$ ($k=1,\ldots,K$) is $N$-order tensor, which represents $\mathcal W(:,...,:,k)$. Denote the $R$ factors CP decomposition of $\mathcal W$ as follows
\begin{align*}
\mathcal W &=\llbracket\mathbf W^{(1)},\mathbf W^{(2)},\ldots,\mathbf W^{(N)},\mathbf W^{(h)}\rrbracket\\
&=\sum_{r=1}^R\mathbf w^{(1)}_{:,r}\circ\mathbf w^{(2)}_{:,r}\circ\cdots\circ\mathbf w^{(N)}_{:,r}\circ\mathbf w^{(h)}_{:,r}.
\end{align*}

Variational EM algorithm is implemented for solving this model. In E-step, we can get $\mathbf u_i$ and $\mathbf \Sigma$
\begin{align}
\mathbf{u}_i = (\mathbf{\Sigma_{\mathcal{W}}}+1/\overline{\rho}\mathbf I_K)^{-1}\mathbf a_i\label{meanh2}
\end{align}
and
\begin{align}
\mathbf \Sigma = (\mathbf I_K+\overline{\rho}\mathbf{\Sigma_{\mathcal{W}}})^{-1},\label{variance2}
\end{align}
where
\[
 \mathbf\Sigma_{\mathcal W}=\mathbf W^{(h)}[(\mathbf W^{(n)T}\mathbf W^{(n)})\odot(\overline{\mathbf W}^{(n)T}\overline{\mathbf W}^{(n)})] \mathbf W^{(h)T},
\]
\[
\mathbf a_{i}=\mathbf W^{(h)}\text{diag}(\mathbf W^{(n)T}\mathbf Y^i_{(n)}\overline{\mathbf W}^{(n)})
\]
and
\[
\overline{\mathbf W}^{(n)}=(\mathbf W^{(N)}\circledast\cdots\circledast\mathbf W^{(n+1)}\circledast\mathbf W^{(n-1)}\circledast\cdots\circledast\mathbf W^{(1)}),
\]

The best $Q(\rho)$ is still a Gamma distribution $\Gamma(\rho|\overline{a},\overline{b})$ with the update parameters:
\begin{align}
\overline{a}= a + \frac{D_1D_2\cdots D_NM}{2}\quad \text{and}\quad\overline{b} = b+\frac{1}{2}\psi\label{rho2}
\end{align}
where
\[
\psi = M \text{tr}(\mathbf\Sigma_{\mathcal{W}}\mathbf\Sigma) +\sum_{i=1}^{M}\|\mathcal Y_i-\mathcal W\times_{(N+1)}\mathbf u_i^T\|_F^2.
\]

In M-step, to solve for such parameterized $\mathcal W$, we gather all the terms related to $\mathcal W$ in (\ref{Exp}) and get
\begin{align*}
\mathcal F^{'}=&\sum_{i=1}^M\|\mathbf Y^i_{(n)}-\mathbf W^{(n)}\text{Diag}(\mathbf u_i^T\mathbf W^{(h)})\overline{\mathbf W}^{(n)T}\|_F^2\\
&+M\text{tr}([(\mathbf W^{(n)T}\mathbf W^{(n)})\odot(\overline{\mathbf W}^{(n)T}\overline{\mathbf W}^{(n)})] \mathbf W^{(h)T} \mathbf\Sigma\mathbf W^{(h)})
\end{align*}
where  $\mathbf Y^i_{(n)}$ represents the mode-$n$ matricized version of $\mathcal Y_i$.

We maximize the lower bound $\mathcal F^{'}$ with respect to factor loadings $\mathbf W^{(n)}$ ($n=1,...,N$) and $\mathbf W^{(h)}$ to obtain:
\begin{align}
\nonumber&\mathbf W^{(n)}\\
\nonumber=&[\sum_{i=1}^M\mathbf Y^i_{(n)}\overline{\mathbf W}^{(n)}\text{Diag}(\mathbf u_i^T\mathbf W^{(h)})]\\
\nonumber&[(\overline{\mathbf W}^{(n)T}\overline{\mathbf W}^{(n)})\odot(\mathbf W^{(h)T} \mathbf\Sigma\mathbf W^{(h)}+\mathbf W^{(h)T}
 \nonumber \mathbf U\mathbf U^T\mathbf W^{(h)})]^{-1}\\
\label{Wn}
\end{align}
and
\begin{align}
\nonumber\mathbf W^{(h)}=&[\mathbf U\mathbf U^T+M\mathbf\Sigma]^{-1}[\sum_i\mathbf u_i[\text{diag}(\overline{\mathbf W}^{(n)T}(\mathbf Y^{i }_{(n)})^T\mathbf W^{(n)})]^T]\\
 &[(\mathbf W^{(n)T}\mathbf W^{(n)})\odot(\overline{\mathbf W}^{(n)T}\overline{\mathbf W}^{(n)})]^{-1}.
\label{Wh2}
\end{align}

The proposed TBV-DR algorithm has time complexity $\mathcal O(RM\prod_{n=1}^ND_n+R^3+K^3)$, which is proven in Supplementary Material. Here $t$ is the actual number of EM iterations, $D_n$ ($n=1,...,N$) are the size of the tensor sample and $M$ is the number of all samples.

High-Order tensor Bayesian vectorial dimension reduction (TBV-DR) algorithm is summarized in Algorithm 2.
\begin{algorithm}
\renewcommand{\algorithmicrequire}{\textbf{Initialize:}}
\renewcommand\algorithmicensure {\textbf{Variational E-step:} }
\renewcommand\algorithmicensure {\textbf{Variational M-step:} }
\caption{High-Order Tensor Bayesian Vectorial Dimension Reduction (TBV-DR) Algorithm}
\begin{algorithmic}[1]
  \REQUIRE Training set $\{\mathcal Y_i, i=1,\ldots,M\}$; Initialize factor matrices $\{\mathbf W^{(n)}\}_{n=1}^N$ and $\mathbf W^{(h)}$, Gamma parameters $a$, $b$ and $\epsilon$.
 \FOR {$t=1$ to $T$}
    %\ENSURE
    \STATE \textbf{Variational E-step:}
    \STATE Maximize the lower bound $\mathcal F$ with respect to $Q(\mathbf h)$: Iterate the variational parameters $\mathbf u_i$ and $\mathbf \Sigma$ in $Q(\mathbf h_i)$ based on (\ref{meanh2}) and (\ref{variance2}) for all $i=1,...,M$.\\
    \STATE Maximize the lower bound $\mathcal F$ with respect to $\rho$: Iterate the variational parameters $a$ and $b$ based on (\ref{rho2})\\
     \STATE \textbf{Variational M-step:}
    \STATE Maximize the lower bound $\mathcal F^{'}$ with respect to $\{\mathbf W^{(n)}\}_{n=1}^N$ and $\mathbf W^{(h)}$: update $\{\mathbf W^{(n)}\}_{n=1}^N$ based on (\ref{Wn}) and update $\mathbf W^{(h)}$ based on (\ref{Wh2}).
        \STATE\textbf{calculate }$e(t)$.
  \STATE\textbf{if } $|e(t)-e(t+1)|<\epsilon$,   break;   \textbf{end}

 \ENDFOR
\end{algorithmic}
\end{algorithm}

\subsection{The Reduced-Dimensionaltiy Representation for a New Sample}
In order to obtain the reduced-dimensionality representation for a given sample, we should solve for the latent variable $\mathbf h_{new}$. From the probabilistic perspective, the posterior mean $\mathbf u_{new}:=\langle\mathbf h_{new}|\mathcal Y_{new}\rangle$ can be seen as the reduced-dimensionality representation, which can be calculated by  (\ref{meanh2}).

\section{Experimental Results and Analysis}\label{Sec:5}
In this section, we conduct some experiments on several publicly available databases to assess the TBV-DR algorithm. These experiments are designed to tell how to select the parameters $R$ and $K$ and demonstrate the performance in recognition and clustering by comparing with several existing models and algorithms. The algorithm is coded in Matlab R2014a and conducted on a PC with a CPU (2.90GHz) and 8G RAMs.

 %As we know, GLRAM is a 2DPCA algorithm. So in order to apply GLRAM on $N$ th-order data tensor with size $D_1\times D_2\times...\times D_N$, we must unfold the tensor to matix. Thus we have the many ways to realize it, such as unfolding matrix with size  $(D_1...D_{i-1}D_{i+1}...D_N)\times D_i$ ($i=1,...,N$). For simple, we restrict to an uniform unfolding matrix with size $(D_1D_2...D_{N-1})\times D_N$.

In the experiments, we set the initial parameters $a =1$, $b=1$, and the factor matrices of all $\mathbf W^{(n)}$ ($n=1,...,N$) and $\mathbf W^{(h)}$ are given randomly. The stopping condition is $\epsilon = 1e^{-4}$ and $T = 200$. We will point it out when these parameters are set to different values.

%We conduct this experiment to test the convergence of TBV-DR algorithm. The experimental data is from the following two databases:
%\begin{itemize}
%  \item A subset of handwritten digits images from the MNIST database (\url{http://yann.lecun.com/exdb/mnist}).
%  \item A light field database of Stanford (\url{http://lightfield.stanford.edu/lfs.html})
%\end{itemize}
\subsection{Experiment 1: Setting $R$ and $K$}\label{expe1}
This experiment aims at illustrating how to select $R$ and $K$. We compare the reconstruction and recognition rates by varying $R$ and $K$.
\begin{table*}[htb]
\renewcommand{\arraystretch}{2.0}
\footnotesize
\centering
\begin{tabular}{ccccccccccccccccccccccc}
  \hline
  \multicolumn{5}{c}{$R=10$}&&\multicolumn{5}{c}{$R=50$}\\
 \cline{1-5} \cline{7-11}
    $K$ & Rate & Std&$e$&Time & &$K$ & Rate& Std&$e$&Time \\
  \hline
 10 &0.3909& 0.0189 &0.7333&4.37 & &30&0.7886&0.0055 &0.7981&13.31\\
20&0.6955&0.0167&0.7491&4.66 &&40&0.8300&0.0046&0.7997&15.22\\
30&0.7053&0.0117&0.7497& 4.93&&50&0.8371&0.0054&0.8006&16.28\\
40&0.7003&0.0168&0.7494& 4.96&&60&0.8419&0.0071&0.8018&16.33\\
50&0.7045&0.0219&0.7516& 5.08&&70&0.8461&0.0079&0.8020&16.47\\
60&0.6946&0.0204&0.7526&  5.68&&80&0.8433&0.0064&0.8022&16.74\\
  \hline
  \multicolumn{5}{c}{$R=100$}&&\multicolumn{5}{c}{$R=200$} \\
 \cline{1-5} \cline{7-11}
  $K$ & Rate & Std&$e$&Time & &$K$ & Rate& Std&$e$&Time\\
  \hline
  70&0.8300&0.0038&0.8356&54.63 & &170&0.8927&0.0020&0.8689&229.53 \\
  80&0.8371&0.0039&0.8362&56.21 &&180&0.8937&0.0024&0.8689&267.15  \\
  90&0.8419&0.0046&0.8364&58.96 &&190&0.8924&0.0016&0.8688&268.25\\
  100&0.8461&0.0040&0.8374&59.69 &&200&0.8915&0.0033&0.8689&284.79\\
  110&0.8433&0.0053&0.8375&62.03 &&210&0.8924&0.0035&0.8690&327.09\\
  120&0.8446&0.0043&0.8376&63.43 &&220&0.8926&0.0027&0.8692&333.46\\
  \hline
\end{tabular}
    \caption{The detailed results on Extended Yale database.}\label{RK}
\end{table*}

The relevant database is the Extended Yale Face (\url{http://vision.ucsd.edu/content/yale-face-database}). This database consists of 2414 frontal-face images of 38 individuals. Each individual has 59 to 64 images. %The images were captured under different illumination and expression conditions.
%All the images are $192\times168$ with 256 grey levels.
Some sample images are shown in Fig. \ref{Yaleface}. In this experiment, all the images are cropped and normalized to a resolution of $32\times 32$ pixels. We use images from all individuals for training and testing. 40 images of each individual are randomly chosen as the training samples, while the remaining images are used for testing. Once features have been extracted by our proposed algorithm, we use the nearest neighbor classifier (1-NN) for classification. Each experiment is run ten times with different random choices. %\GaoC{It seems you did not talk about how to infer $h$ for new testing data after $W$ has been learned?  Or do you assume readers know about this?}

In this experiment, we recorded the average recognition rates and variances on testing sample set, the reconstruction errors $e$ in (\ref{define_e}) on training sample set, and the time consumption in TABLE \ref{RK}. In this table, we list these results with four different $R$'s (10, 50, 100 and 200). After fixing $R$, we test the above algorithm by varying the value of $K$. From this table, we can observe that
% * <yfsun@bjut.edu.cn> 2016-09-18T01:48:37.525Z:
%
% ^.
% * <yfsun@bjut.edu.cn> 2016-09-18T01:48:21.656Z:
%
% ^.
% * <yfsun@bjut.edu.cn> 2016-09-18T01:48:21.563Z:
%
% ^.
% * <yfsun@bjut.edu.cn> 2016-09-18T01:48:21.272Z:
%
% ^.
% * <yfsun@bjut.edu.cn> 2016-09-18T01:45:10.124Z:
%
% ^.
\begin{itemize}
\item The recognition rates are significantly improved and the variances are reduced obviously with increasing $R$. It also illustrates the proposed algorithm is more stable. We note that the value of $R$ should not be too small.
\item When the $R$ is fixed, the recognition rates tends to increase with increasing $K$. However, for smaller $K$ values, the recognition rate can be influenced. When $K$ is much bigger than $R$, the recognition rates may reduce. We believe this is due to the overfitting.
\item When fixing $K$, the recognition rates tends to increase with increasing $R$. For example, the recognition rate grows from 0.7003 to 0.8300 for $R=10$, $K=40$ to $R=50$, $K=40$. When $R$ is much bigger than $K$, the recognition rates may reduce. For example, the recognition rate reduces from 0.8461 to 0.8300 for $R=50$, $K=70$ to $R=100$, $K=70$.
\end{itemize}
From the above analysis, we can come up with two principles for setting $R$ and $K$. First, $R$ should not be too small. Second, the gap between $K$ and $R$ shouldn't be too large. For convenience, we can set $K=R$ in investigating recognition rate, time consuming and overfitting. %\GaoC{If $R=K$, then it would become CP model?? Also I am curious why you come to $R=K$.  If your examples above, have you demonstrated things are getting worse when $K-R$ is getting large. Otherwise how do you conclude that it is better for $K=R$} However, when the value of $R$ is vary big, reducing the value of $K$ appropriately also have good results. For example $R=200$ and $170 \leq K\leq 200$. But for clearness, we set $R=K$ in the following experiments.

\subsection{Experiment 2: Face Recognition}
All of the face recognition experiments are conducted on three publicly available databases:
\begin{itemize}
  \item The extended Yale face database, which is same as that introduced in Section \ref{expe1}%\url{http://vision.ucsd.edu/content/yale-face-database}.
  %\item The ORL face database \url{http://www.cl.cam.ac.uk/research/dtg/attarchive/facedatabase.html}.
  \item The AR face database \url{http://rvl1.ecn.purdue.edu/aleix/aleix_face_DB.html}.
  \item The FERET face database \url{http://www.itl.nist.gov/iad/humanid/feret/feret_master.html}.
\end{itemize}

\begin{figure}
\begin{center}
{\includegraphics[width=90mm,height=30mm]{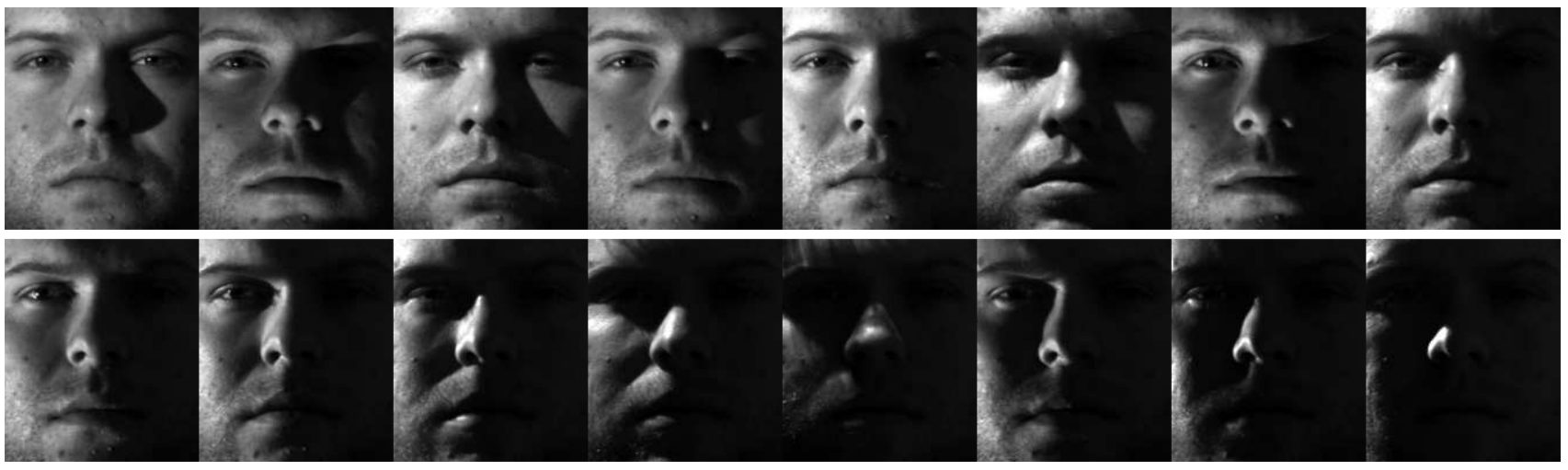}}
\end{center}
\caption{Some sample images from the extended Yale database.}\label{Yaleface}
\end{figure}
%The ORL face database includes 40 distinct individuals with 10 images for each individual. For some individuals, the images were taken at different times, illumination changes, facial expressions variation (open / closed eyes, smiling / not smiling) and facial details (glasses / no glasses). All images are grayscale and normalized to a resolution of $64\times 64$ pixels.
  \begin{figure*}
\begin{center}
{\includegraphics[width=170mm,height=45mm]{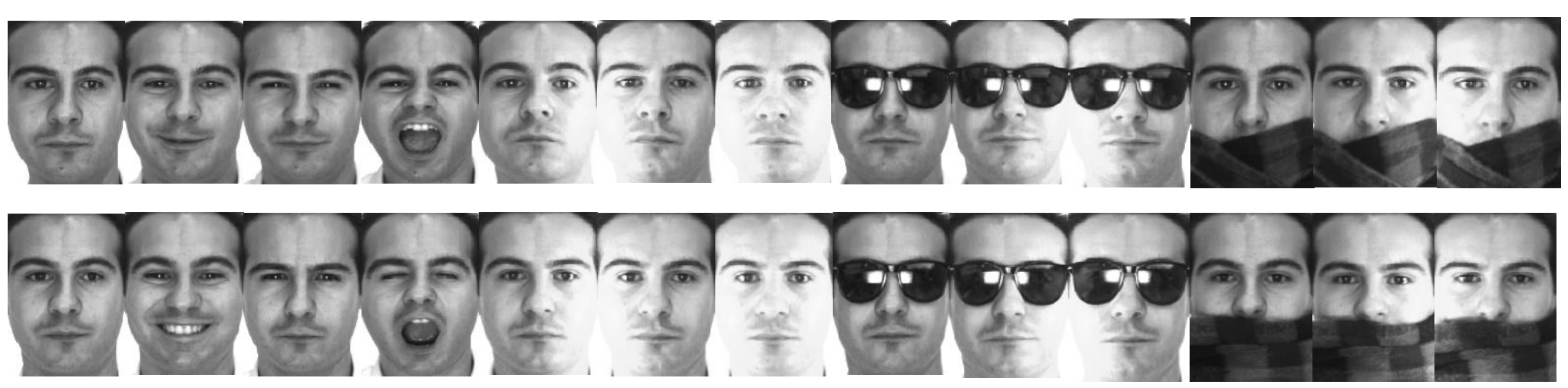}}
\end{center}
\caption{Twenty-six face examples of one subject from AR database. The first row images are from the first session, and the second row is from the second session.}\label{ARface}
\end{figure*}

The AR face database contains over 4,000 color images corresponding to 126 subjects. There are variations of facial expressions, illumination conditions and occlusions (sunglasses and scarf) for  each subject. Each individual has 26 frontal view images taken in two sessions (separated by 2 weeks), where in each session there are 13 images. The original images are of size $768 \times 576$ pixels and of 24 bits of RGB color values. Fig. \ref{ARface} shows 26 images of one individual. All images are cropped and resized to $50\times 40$ pixels.

FERET database includes 1400 images of 200 different subjects, with 7 images per subject. All images  are grayscale and scaled to a resolution of $32\times 32$ pixels. Some sample images are shown in Fig. \ref{feret}.
\begin{figure}
\begin{center}
{\includegraphics[width=85mm,height=14mm]{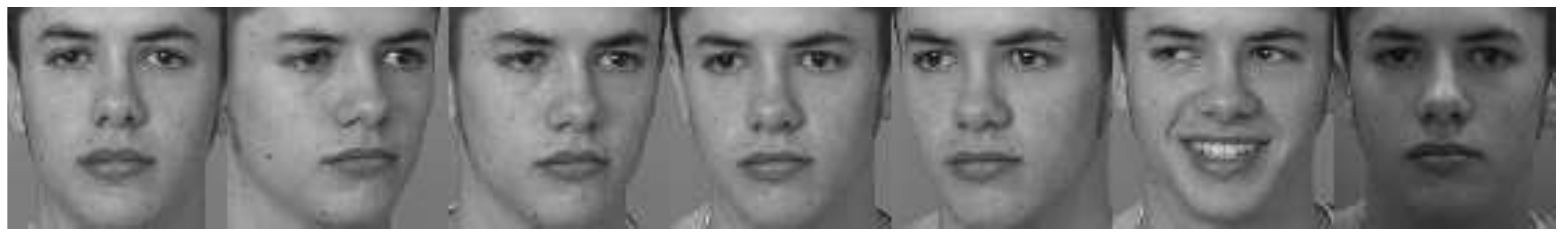}}
\end{center}
\caption{All images of one individual from the FERET database.}\label{feret}
\end{figure}

\begin{table*}
\renewcommand{\arraystretch}{1.3}
\footnotesize
\centering
\begin{tabular}{ccccccc}
  \hline
\multirow{2}*{Images}&\multirow{2}*{Algorithm}&\multicolumn{5}{c}{feature number}\\
 \cline{3-7}
     &  & 50&70&100&200&300 \\
  \hline
\multirow{5}*{20} &PCA& 0.4522&0.4891 &0.5133&0.5472&0.5562\\
&mixPPCA&0.6973$\pm$0.0242&0.6918$\pm$0.0146&0.6966$\pm$0.0198&0.6903$\pm$0.0371&0.6953$\pm$0.0397\\
&GLRAM ($r,c$)&0.4687 (8,8)&0.5045 (9,9)&0.5246 (10,10)&0.6353 (15,15)&0.6588 (18,18)\\
&CP&0.6242$\pm$ 0.0155&0.6840$\pm$0.0149&0.7406$\pm$0.0157&0.7866$\pm$0.0067&0.7678$\pm$0.0065\\
&Bayesian CP&0.5042&0.6487&0.7412&0.8065&0.8138\\
&TBV-DR&\textbf{0.7941}$\pm$\textbf{0.0053}&\textbf{0.8105}$\pm$\textbf{0.0037}&\textbf{0.8268}$\pm$\textbf{0.0034}&\textbf{0.8489}$\pm$\textbf{0.0011}&\textbf{0.8489}$\pm$\textbf{0.0011}\\
  \hline
 \multirow{5}*{30} &PCA& 0.5078&0.5518 &0.5840 &0.6287&0.6342\\
&mixPPCA&0.7365$\pm$0.0130&0.7684$\pm$0.0081&0.7733$\pm$0.0219&0.7807$\pm$0.0197&0.7802
$\pm$0.0189
\\
&GLRAM ($r,c$) &0.4199 (8,8)&0.4521 (9,9)&0.4765 (10,10)&0.5730 (15,15)&0.5997 (18,18)\\
&CP&0.6713$\pm$0.0187&0.7217$\pm$0.010&0.7762$\pm$0.0125&0.8328$\pm$0.0064&0.8242$\pm$0.0062\\
&Bayesian CP&0.5479&0.6954&0.7755&0.8611&0.8485\\
&TBV-DR&\textbf{0.8248}$\pm$\textbf{0.0042}&\textbf{0.8458}$\pm$\textbf{0.0060}&\textbf{0.8627}$\pm$\textbf{0.0054}&\textbf{0.8856}$\pm$\textbf{0.0021}&\textbf{0.8893}$\pm$\textbf{0.0008}\\
  \hline
 \multirow{5}*{40} &PCA& 0.5671 & 0.6096&0.6643&0.6890&0.6957\\
&mixPPCA&0.7800$\pm$0.0075&0.7918$\pm$0.0087& 0.8093
$\pm$0.0102&0.8103$\pm$0.0208&0.8087$\pm$0.0125\\
&GLRAM ($r,c$)&0.3501 (8,8)&0.3845 (9,9)&0.4027 (10,10)&0.4940 (15,15)&0.5206 (18,18)\\
&CP&0.6242$\pm$0.0155&0.6840$\pm$0.0149&0.7406$\pm$0.0157&0.7866$\pm$0.0067&0.7678$\pm$0.0065\\
&Bayesian CP&0.5861&0.7383&0.7953&0.8624&0.8747\\
&TBV-DR&\textbf{0.8371}$\pm$\textbf{0.0054}&\textbf{0.8578}$\pm$\textbf{0.0042}&\textbf{0.8749}$\pm$\textbf{0.0040}&\textbf{0.8915}$\pm$\textbf{0.0033}&\textbf{0.9028}$\pm$\textbf{0.0027}\\
  \hline
\end{tabular}
    \caption{Recognition rates and variances of standard PCA, mixPPCA, GLRAM, CP and TBV-DR training on extended Yale database.}\label{table_Yale}
\end{table*}
\begin{table*}[htb]
\renewcommand{\arraystretch}{1.3}
\footnotesize
\centering
\begin{tabular}{ccccccc}
  \hline
\multirow{2}*{Individual}&\multirow{2}*{Algorithm}&\multicolumn{5}{c}{feature number}\\
 \cline{3-7}
     &  & 50&70&100&200&300 \\
  \hline
\multirow{5}*{30} &PCA& 0.7762&0.7810 &0.7810&0.7810&0.7810\\
&mixPPCA&0.7614$\pm$0.0220(10)&0.8314$\pm$0.0643(20)&0.8286$\pm$0.0452(30)&0.7043$\pm$0.2757(40)&0.6490$\pm$0.2890(50)\\
&GLRAM ($r,c$)&0.7190 (8,8)&0.7286 (9,9)&0.7381 (10,10)&0.7619 (15,15) &0.7714 (18,18)\\
&CP&0.7152$\pm$ 0.0248&0.7452$\pm$0.0185&0.7386$\pm$0.0233&0.7252$\pm$0.0209&0.6838$\pm$0.0131\\
&Bayesian CP&0.6048&0.7571&0.7619&0.7667&0.8238\\
&TBV-DR&\textbf{0.8824}$\pm$\textbf{0.0095}&\textbf{0.8895}$\pm$\textbf{0.0070}&\textbf{0.8943}$\pm$\textbf{0.011}&\textbf{0.8867}$\pm$\textbf{0.0070}&\textbf{0.8757}$\pm$\textbf{0.0069}\\
  \hline
 \multirow{5}*{50} &PCA&0.7229 &0.7314 &0.7486 &0.7514&0.7571\\
&mixPPCA&0.6134$\pm$0.0736(10)&0.7189$\pm$0.0231(20)&0.7383$\pm$0.0296(30)&0.7949$\pm$0.0272(40)&0.7003$\pm$0.1813(50)\\
&GLRAM ($r,c$)&0.6714 (8,8)&0.6857 (9,9)&0.7029 (10,10)&0.7486 (15,15)&0.7514 (18,18)\\
&CP&0.6371$\pm$0.0200&0.6557$\pm$0.0202&0.6666$\pm$0.0158&0.6517$\pm$0.0118&0.6371$\pm$0.0138\\
&Bayesian CP&0.5514&0.5717&0.6600&0.7171&0.6914\\
  &TBV-DR&\textbf{0.8377}$\pm$\textbf{0.0085}&\textbf{0.8457}$\pm$\textbf{0.0093}&\textbf{0.8403}$\pm$\textbf{0.0088}&\textbf{0.8183}$\pm$\textbf{0.0082}&\textbf{0.8051}$\pm$\textbf{0.0087}\\
  \hline
 \multirow{5}*{70} &PCA& 0.7265 &0.7490 &0.7735&0.7816&0.7796\\
&mixPPCA&0.6086$\pm$0.0506(10)&0.7088$\pm$0.0194(20)&0.7365$\pm$0.0268(30) &0.7714$\pm$0.0138 &0.7839$\pm$0.015(50)\\
&GLRAM ($r,c$)&0.6592 (8,8)&0.6878 (9,9) &0.7102 (10,10)&0.7388 (15,15)&0.7592 (18,18)\\
&CP&0.6082$\pm$0.0214&0.6253$\pm$0.0255&0.6422$\pm$0.0216&0.6380$\pm$0.0195&0.5939$\pm$0.0216\\
&Bayesian CP&0.4898&0.5796&0.6061&0.6694&0.7020\\
&TBV-DR&\textbf{0.8061}$\pm$\textbf{0.0137}&\textbf{0.8106}$\pm$\textbf{0.0112}&\textbf{0.8224}$\pm$\textbf{0.0060}&\textbf{0.8382}$\pm$\textbf{0.0085}&\textbf{0.8347}$\pm$\textbf{0.0053}\\
  \hline
\end{tabular}
    \caption{Recognition rates of standard PCA, mixPPCA, GLRAM, CP and TBV-DR training on AR database.}\label{table_AR}
\end{table*}
In this experiment, we compare the TBV-DR algorithm against the standard PCA,  mixture of PPCA (mixPPCA)} \cite{TippingBishop1999},
GLRAM (Generalized Low Rank Approximations of Matrices) \cite{Ye2005}, CP decomposition, Bayesian CP \cite{ZhaoZhouZhangCichockiAmari2016} and Tucker-2 decomposition \cite{KoldaBader2009}. As the Tucker-2 model with HOOI algorithm \cite{KoldaBader2009} is equivalent to GLRAM \cite{YuBiYe2008}, we only compare our model with GLRAM. In the CP-based algorithms of all ($N+1$)-order tensor samples $\mathcal Y\in \mathbb R^{D_1\times...\times D_N\times M}$, the final mode factor matrix $\mathbf W\in \mathbb R^{M\times R}$ can be used as the latent CP features while other mode matrices can be considered as the basis of latent subspace. The number of features is determined by $R$, which is also the factors number in CP decomposition. The code of the Bayesian CP decomposition  can be downloaded from \url{http://www.bsp.brain.riken.jp/~qibin/homepage/Software.html.}

Thus we can extract features for each sample in the training set by each of the algorithms, and then use the nearest neighbor classifier (1-NN) for classification. All experiments are run ten times, and the average recognition rates and variance are recorded.
\begin{comment}
\begin{table}[htb]
\renewcommand{\arraystretch}{2.0}
\footnotesize
\centering
\begin{tabular}{|c|c|c|c|c|c|}
  \hline
  Algorithm & 20 &30 & 40\\
  \hline
Standard PCA & 0.5278& 0.6005& 0.6745 \\
    \hline
mixPPCA  &0.6446$\pm$0.0341 & 0.7131$\pm$0.0114&0.7557$\pm$0.0130   \\
  \hline
GLRAM &0.5677 & 0.6515 & 0.7069\\
  \hline
TBV-DR &\textbf{0.7322}$\pm$\textbf{0.0039}& \textbf{0.7672}$\pm$\textbf{0.0050} & \textbf{0.7812}$\pm$\textbf{0.0083}\\
  \hline
\end{tabular}
    \caption{Recognition rate of standard PCA, mixPPCA, GLRAM and TBV-DR training on extended Yale database.}\label{table_Yale}
\end{table}
\end{comment}
\begin{table*}[htb]
\renewcommand{\arraystretch}{1.3}
\footnotesize
\centering
\begin{tabular}{ccccccc}
  \hline
\multirow{2}*{Individual}&\multirow{2}*{Algorithm}&\multicolumn{5}{c}{feature number}\\
 \cline{3-7}
     &  & 50&70&100&200&300 \\
  \hline
\multirow{5}*{30} &PCA&0.4000&0.4167 &0.4667&0.4833&0.4833\\
&mixPPCA&0.4833$\pm$0.0401(10)&0.4900$\pm$0.0522(20)&0.3350$\pm$0.2226(30)&0.3383$\pm$0.2571(40)&0.3266$\pm$0.2069(50)\\
&GLRAM ($r,c$)&0.4833 (8,8)&0.4833 (9,9)&0.5000 (10,10)&0.5000 (15,15)&0.5167 (18,18)\\
&CP&0.4900$\pm$ 0.0417&0.5283$\pm$0.0614&0.4800$\pm$0.0637&0.3567$\pm$0.0387&0.3017$\pm$0.0277\\
&Bayesian CP&0.4167&0.5000&0.5667&0.5833&0.5667\\
&TBV-DR&\textbf{0.7433}$\pm$\textbf{0.0370}&\textbf{0.7783}$\pm$\textbf{0.0223}&\textbf{0.7983}$\pm$\textbf{0.0299}&\textbf{0.7833}$\pm$\textbf{0.0136}&\textbf{0.7583}$\pm$\textbf{0.0196}\\
  \hline
 \multirow{5}*{50} &PCA&0.4200 &0.4300 &0.4400&0.4600&0.4600\\
&mixPPCA&0.57300$\pm$0.0211(10)&0.6060$\pm$0.0331(20)&0.6460$\pm$0.0558(30)&0.5120$\pm$0.2271&0.5360$\pm$0.2678(50)
\\
&GLRAM ($r,c$)&0.5400 (8,8)&0.5400 (9,9)& 0.5500 (10,10)&0.5600 (15,15)& 0.5600 (18,18)\\
&CP&0.4910$\pm$0.0555&0.4760$\pm$0.0433&0.5220$\pm$0.0627&0.4450$\pm$0.0472&0.3740$\pm$0.0255\\
&Bayesian CP&0.3600&0.4300&0.4600&0.5200&0.5100\\
&TBV-DR&\textbf{0.7820}$\pm$\textbf{0.0294}&\textbf{0.7950}$\pm$\textbf{0.0212}&\textbf{0.7960}$\pm$\textbf{0.0291}&\textbf{0.7840}$\pm$\textbf{0.0165}&\textbf{0.7590}$\pm$\textbf{0.0228}\\
  \hline
 \multirow{5}*{100} &PCA&0.3100 &0.3450 &0.3650&0.3850&0.3900\\
&mixPPCA&0.4680$\pm$0.0212(10)&0.5655$\pm$0.0196(20)&0.5975$\pm$0.0221(30)&0.6055$\pm$0.0244(40)&0.6205$\pm$0.0248(50)\\
&GLRAM ($r,c$)&0.4850 (8,8)&0.4800 (9,9)&0.4850 (10,10)&0.4850 (15,15)&0.4800 (18,18)\\
&CP&0.3920$\pm$0.0544&0.4105$\pm$0.0487&0.4260$\pm$0.0336&0.3580$\pm$0.0220&0.2740$\pm$0.0247\\
&Bayesian CP&0.3050&0.3700&0.4100&0.4200&0.4150\\
&TBV-DR&\textbf{0.6915}$\pm$\textbf{0.0227}&\textbf{0.6770}$\pm$\textbf{0.0138}&\textbf{0.6675}$\pm$\textbf{0.0177}&\textbf{0.7020}$\pm$\textbf{0.0082}&\textbf{0.6705}$\pm$\textbf{0.0086}\\
  \hline
\end{tabular}
    \caption{Recognition rates of standard PCA, mixPPCA, GLRAM, CP and TBV-DR training on FERET database.}\label{table_FERET}
\end{table*}
\begin{table}[htb]
\renewcommand{\arraystretch}{1.3}
\footnotesize
\centering
\begin{tabular}{ccccccc}
  \hline
\multirow{2}*{Sample}&\multirow{2}*{Algorithm}&\multicolumn{5}{c}{feature number}\\
 \cline{3-7}
     &  & 50&70&100&200&300 \\
  \hline
\multirow{5}*{Yale (20)} &PCA&1.73&1.79&2.19&3.3&3.89\\
&mixPPCA&24.52&33.10&44.73&93.11&211.79\\
&GLRAM&1.58&1.62&1.78&2.82&3.68\\
&CP&8.38&8.70&13.44&24.91&36.26\\&Bayesian CP&4.42&9.55&12.34&83.84&120.25\\
&TBV-DR&8.93&13.40&22.01&72.23&157.68\\
  \hline
\multirow{5}*{AR (30)} &PCA&0.22&0.23&0.24&0.26&0.31\\
 &mixPPCA&2.39&2.76&2.89&2.94&3.29\\
&GLRAM&0.11&0.12&0.13&0.19&0.26\\
&CP&3.06&5.11&7.33&12.73&17.80\\
&Bayesian CP&1.17&2.20&11.22&16.83&22.04\\
&TBV-DR&2.40&4.69&7.44&26.87&86.56\\
  \hline
   \multirow{5}*{FERET (30)} &PCA&0.08&0.09&0.09&0.10&0.11\\
   &mixPPCA&0.63&0.76&1.03&2.09&3.02\\
&GLRAM&0.04&0.05&0.06&0.06&0.06\\
&CP&2.83&3.51&5.47&9.68&13.26\\
&Bayesian CP&2.04&2.61&4.13&7.50&20.20\\
&TBV-DR&2.22&3.17&6.36&25.15&53.85\\
  \hline
\end{tabular}
    \caption{Time consuming of standard PCA, mixPPCA, GLRAM, CP and TBV-DR training on Yale, AR and FERET databases.}\label{timeTable}
\end{table}

\textbf{Extended Yale Face Database:}
In the experiment on extended Yale database, we design three tests to compare recognition rates of the aforementioned algorithms. In these tests, we randomly select 20, 30, 40 images of each individual as the training samples, respectively. The remaining images are used for testing. We record the recognition rates with the feature number $K$ being 50, 70, 100, 200 and 300 in every test. The average recognition rates and variances from ten runs are listed in TABLE \ref{table_Yale}. As GLRAM is a 2DPCA algorithm, we list its recognition rates as well as the reduced dimensionality ($r,c$) in the table.

From the table we can see that TBV-DR algorithm achieves the highest recognition rates with smaller variances. The highest recognition rates of PCA on 20, 30, 40 training samples are 0.5617, 0.6374 and 0.7047, respectively. In GLRAM algorithm, we get the highest recognition rates of 0.5617, 0.6421 and 0.7013,  respectively. This indicates that our model can outperform PCA and GLRAM (or Tucker-2) in terms of recognition. In Bayesian CP decomposition,
 we apply SVD to the data and initialize factor matrices. Thus the recognition rates of ten times are not vary. In addition, comparing with CP and mixture PPCA, we also get the best classification performance for our algorithm. In the CP algorithm, both the subspace basis and feature number are determined by $R$. Raising $R$ will result in overfitting, such as $K=300$. However, in our model, the number of features is determined by $K$, while the flexibility of subspace basis is determined by $R$. This implies that we can obtain very different bases even the number of features is fixed. For example, the recognition rates become from 0.8371 ($R=50$) to 0.8452 ($R=100$) when $K=50$. Therefore, the proposed TBV-DR model might provide a flexible and effective feature extraction framework.
\textbf{AR Face Database:}
In this experiment, we randomly select 30 (15 men and 15 women), 50 (25 men and 25 women) and 70  individuals (50 men and 20 women) respectively to test and only use the non-occluded 14 images (i.e., the first seven face images of each row in Fig. \ref{ARface}). The first seven of each individual are used for training and the last seven for testing. The average recognition rates and variances across ten rounds of experiments are shown in TABLE \ref{table_AR}.

In mixture of PPCA algorithm, the recognition rates drop rapidly when the feature number exceed 50. So we reduce the feature number to less than 50 and record the recognition rates. The numbers in brackets represent the corresponding feature number, i.e., the number of reduced dimensionaltiy. From this table, we can see that mixture of PPCA algorithm is greatly affected by the feature number. For 30 individuals as an example,  the recognition rates drop from 0.8286 to 0.7043 when we increase the reduced dimensionaltiy from 30 to 40.
Compared with all the other algorithms, the proposed TBV-DR algorithm can achieve the overall best recognition performance.
\begin{comment}
\begin{table}[htb]
\renewcommand{\arraystretch}{2.0}
\footnotesize
\centering
\begin{tabular}{|c|c|c|c|c|c|}
  \hline
 Algorithm & 30 &50 & 70\\
  \hline
Standard PCA & 0.7810& 0.7514& 0.7796 \\
    \hline
mixPPCA  & 0.8619$\pm$0.0304& 0.7837$\pm$0.0173&0.7867$\pm$0.0233 \\
  \hline
 GLRAM &0.7762 &0.7600  & 0.7796 \\
  \hline
TBV-DR &\textbf{0.8786}$\pm$\textbf{0.0192} &\textbf{0.8397}$\pm$\textbf{0.0120} &\textbf{0.7892}$\pm$\textbf{0.0093} \\
  \hline
\end{tabular}
    \caption{Recognition rate of standard PCA, mixPPCA, GLRAM and TBV-DR training on AR face database.}\label{table_AR}
\end{table}
\end{comment}
 \begin{table*}
\renewcommand{\arraystretch}{1.3}
\footnotesize
\centering
\begin{tabular}{ccccccc}
  \hline
 \multirow{2}*{Algorithm}& Evaluation& \multicolumn{3}{c}{feature number}\\
  \cline{3-5}
  & metric&10&30&50\\
    \hline
\multirow{2}*{LRR} & AC &0.5344$\pm$0.0047&0.5077$\pm$0.0063&0.4964$\pm$0.0105\\

  & NMI & 0.5184$\pm$0.0038&0.4902$\pm$0.0090 &0.5082$\pm$0.0297 \\
    \hline
\multirow{2}*{SSC}&AC &0.5645$\pm$0.0032&0.5074$\pm$0.0060&0.5018$\pm$0.0124\\
  & NMI & 0.5193$\pm$0.0042&0.4918$\pm$0.0035 &0.5113$\pm$0.0369 \\
    \hline
\multirow{2}*{PCA+Kmeans}&AC& \textbf{0.5780}$\pm$\textbf{0.0337}&0.5600$\pm$0.0609&0.5249$\pm$0.0745\\
  & NMI & \textbf{0.5597}$\pm$\textbf{0.0265}&0.5207$\pm$0.0501 &0.5129$\pm$0.0586 \\

  \hline
\multirow{2}*{CP+Kmeans}&AC&0.5611$\pm$0.0507&0.5593$\pm$0.0349&0.5408$\pm$0.0570 \\
  & NMI & 0.5449$\pm$0.0471&0.5406$\pm$0.0320 &0.5209$\pm$0.0537 \\
  \hline
 \multirow{2}*{ Bayesian CP+Kmeans}&AC &0.5885$\pm$0.0326 &0.5206$\pm$0.0649&0.4950$\pm$0.0566 \\
   & NMI & 0.5280$\pm$0.0245& 0.5195$\pm$0.0533& 0.5090$\pm$0.0515\\
  \hline
 \multirow{2}*{ TUCKER+Kmeans}&AC  &0.5755$\pm$0.0384&0.4677$\pm$0.0545&0.3979$\pm$0.0587 \\
    & NMI &0.5337$\pm$0.0424 &0.4019$\pm$0.0670 &0.3345$\pm$0.0604 \\
  \hline
 \multirow{2}*{TBV-DR+Kmeans}&AC &0.5593$\pm$0.0445& \textbf{0.5616}$\pm$\textbf{0.0416}&\textbf{0.5889}$\pm$\textbf{0.0357}\\
  & NMI &0.5379$\pm$0.0362 &\textbf{0.5543}$\pm$\textbf{0.0360} &\textbf{0.5691}$\pm$\textbf{0.0395} \\
  \hline
\end{tabular}
    \caption{Subspace clustering results for different algorithms on the Ballet database}.\label{Ballet}
\end{table*}

\textbf{FERET Face Database:} FERET database contains 200 individuals with 7 images per individual. In this experiment, we randomly select 30, 50, 100 individuals respectively to test the recognition rates. 5 images of each individual who is selected by us are used for training and 2 remaining images are used for testing. The average recognition rates are shown in TABLE \ref{table_FERET}.

Like above illustrations, we list the classification results of GLRAM and mixture PPCA as well as their feature numbers. From the table, we can see the consistent results with the experimental results on AR database.
\begin{comment}
\begin{table}[htb]
\renewcommand{\arraystretch}{2.0}
\footnotesize
\centering
\begin{tabular}{|c|c|c|c|c|c|}
  \hline
  Algorithm & 30 &50& 100\\
  \hline
Standard PCA & 0.4667& 0.4600& 0.3750 \\
    \hline
mixPPCA &0.5417$\pm$0.0126 & 0.6290$\pm$0.0387&0.5845$\pm$0.0336  \\
  \hline
GLRAM &0.5167 & 0.5600 & 0.4950 \\
  \hline
TBV-DR &\textbf{0.7617}$\pm$\textbf{0.0294} & \textbf{0.6790}$\pm$\textbf{0.0251}&\textbf{0.6480}$\pm$\textbf{0.0155} \\
  \hline
\end{tabular}
    \caption{Recognition rate of standard PCA, mixPPCA, GLRAM and TBV-DR training on FERET face database .}\label{table_FERET}
\end{table}
\end{comment}

From TABLEs \ref{table_Yale}, \ref{table_AR} and \ref{table_FERET}, we can observe that TBV-DR can get much better recognition results with both fewer training samples or more categories. Thus we can directly use TBV-DR algorithm to extract feature of 2D data and get effective representation.

To illustrate the proposed algorithm complexity, we report the time consumption of all the above algorithms in TABLE \ref{timeTable}. ``Yale (20)" represents that we randomly select 20 images of each individual as the training samples as in TABLE \ref{table_Yale}. ``AR (30)" and ``FERET (30)'' are corresponding to TABLE \ref{table_AR} and \ref{table_FERET}, respectively. The time consumption of mixPPCA has reduced significantly from ``Yale (20)'' to ``AR (30)'' and ``FERET (30)'', that's because of the different feature numbers. The least time consumption is PCA and GLRAM. Although our algorithm consumes the most time, in order to get higher recognition rate, it is also acceptable with moderate $K$.

\subsection{Experiment 3: Clustering}
In this experiment, we conduct clustering third-order tensor data on the following publicly available database:
\begin{itemize}
  \item Ballet Action Database \cite{FathiMori2008} \url{https://www.cs.sfu.ca/research/groups/VML/semilatent/}.
\end{itemize}

\begin{figure}
\begin{center} {\includegraphics[width=85mm,height=35mm]{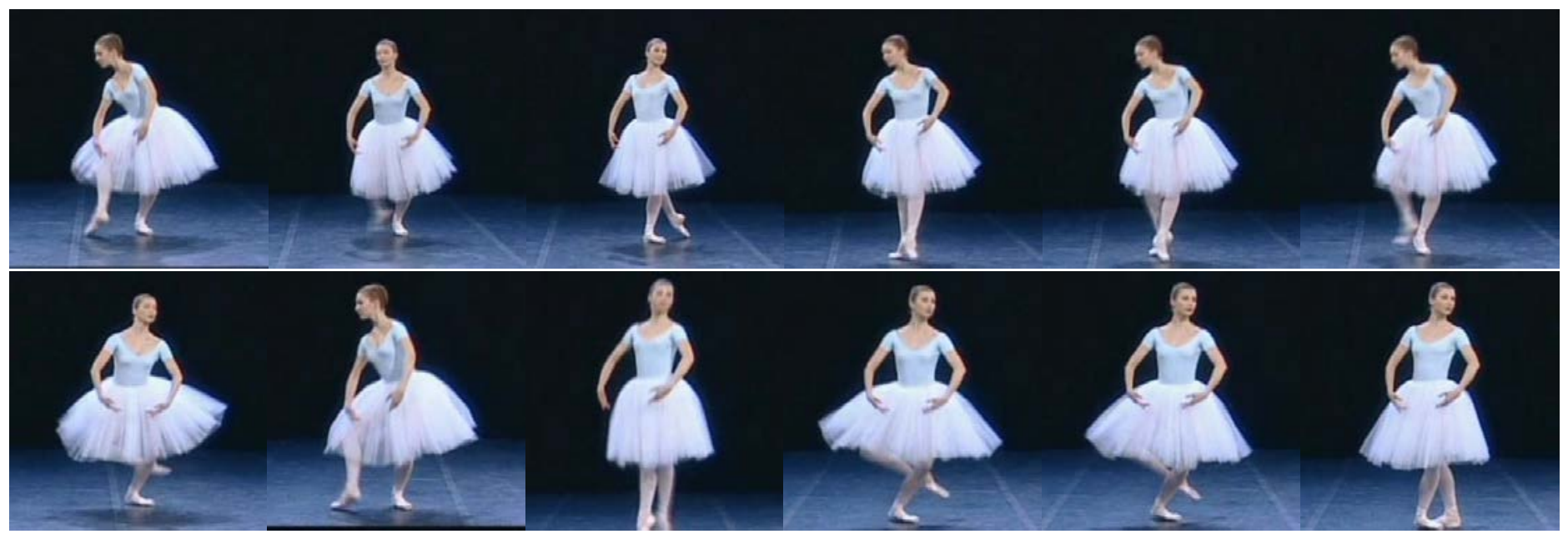}}
\end{center}
\caption{Some samples from the Ballet Action database.}\label{BalletDataset}
\end{figure}

Ballet Action Database contains 44 video clips, collected from an instructional ballet DVD. The database consists of 8 complex action patterns performed by 3 subjects. These 8 actions include: `left-to-right hand opening', `right-to-left hand opening', `standing hand opening', `leg swinging', `jumping', `turning', `hopping' and `standing still'. Clustering this database is challenging due to the significant intra-class variations in terms of speed, spatial and temporal scale, clothing and movement. The frame images are normalized and centered in a fixed size of $30 \times 30$. Some frame samples of Ballet database are shown in Fig. \ref{BalletDataset}.

In this experiment, we split each clip into subgroups of 12 images and each subgroup is treated as an image set. As a result, we construct a total of $713$ image sets which are labeled with $8$ clusters. This database which does not contain complex background or illumination changes can be regarded as a clean human action data in ideal condition.

  %%According to the above data preparation, we can get the data $\mathcal Y\in \mathbb R^{30\times 30\times12 \times 713}$, which is 4D tensor data containing the collection of 713 3D tensors with size $30\times 30\times 12$.
In this experiment, the TBV-DR algorithm is compared against PCA, CP, Bayesian CP, Tucker decomposition algorithm and two other popular cluster algorithms: Low Rank Representation (LRR) \cite{LiuLinSunYuMa2013} and Sparse Subspace Clustering (SSC) \cite{ElhamifarVidal2013}. In Tucker decomposition, we need to find subspace $\mathbf V_{1}\in \mathbb R^{30\times i_1}$, $\mathbf V_{2}\in \mathbb R^{30\times i_2}$, $\mathbf V_{3}\in \mathbb R^{12\times i_3}$, $\mathbf V_{4}\in \mathbb R^{713\times i_4}$ and core tensor $\mathbf C\in \mathbb R^{i_1\times i_2\times i_3\times i_4}$. Thus the optimal $\mathbf V_{4}$  can be regarded as the low-dimensional embedding for data $\mathcal Y$ and cluster feature \cite{SunGaoHongMishraYin2015}. The PCA, CP, Bayesian CP, Tucker and TBV-DR are algorithms based on dimensionality reduction. For these algorithms, after extracting features, the k-means can be used for clustering. LRR and SSC are two clustering methods building on the subspace clustering framework. We  mainly use them as references for performance assessment.

To quantitatively evaluate the clustering results, we adopt two evaluation metrics, the \emph{accuracy} (AC) and the \emph{normalized mutual information} (NMI) metrics, more materials about AC and NMI can be referred to \cite{YinGaoLin2015}. We run each algorithm ten times for each dataset and record the average and variance of AC and NMI.
  \begin{comment}
\begin{table}[htb]
\renewcommand{\arraystretch}{1.5}
\small
\centering
\begin{tabular}{|c|c|c|c|c|c|}
  \hline
 Algorithm & Parameter setting\\
  \hline
Standard PCA &$d=130$   \\
  \hline
TUCKER & $i_1=15$, $i_2=15$, $i_3=6$, $i_4=10$\\
  \hline
TBV-DR & $K = 45$, $R=40$\\
  \hline
\end{tabular}
    \caption{Parameter setting for different algorithms.}\label{parameter}
\end{table}
\end{comment}

TABLE \ref{Ballet} presents the experimental results of all the algorithms on the Ballet database. However, the results in this table are not very high for 8 clusters. The reason is that some kind of actions are too similar to distinguish, i.e. `left-to-right hand opening' and `right-to-left hand opening'. From the table, we can see that the proposed TBV-DR algorithm is comparable with or even better than other algorithms.
\subsection{Experiments 4: Text Classification and Data Reconstruction}
In this section, we tend to explore how the proposed method behaves on any data besides images/audio/videos. The related dataset is
\begin{itemize}
\item Berkeley dataset (\url{http://db.csail.mit.edu/labdata/labdata.html})
\item IMDB text dataset (\url{http://ai.stanford.edu/~amaas/data/sentiment/})
\end{itemize}

\textbf{Berkeley dataset}: this dataset contains information about data collected from 54 sensors deployed in the Intel Berkeley Research lab between February 28th and April 5th, 2004.
Mica2Dot sensors with weather boards collected timestamped topology information, along with humidity, temperature, light and voltage values once every 31 seconds. Data was collected using the TinyDB in-network query processing system, built on the TinyOS platform.

As collected data at 2 sensing nodes are missing too much, so we choose temperature, humidity, light and voltage data at 52 sensing nodes and 2880 time nodes in this experiment. Thus we can obtain a tensor $\mathcal X$ with size of $52\times 2880\times 4$. In PCA, we reshape $\mathcal X$ to a matrix with size of $52 \times (2880*4)$ and 52 represents samples number. However, this matrix is a low rank matrix and PCA can not work on it. Thus in this dataset, we only test the reconstruction performance of our proposed method against CP decomposition, in terms of the following relative reconstruction error.  %All data are normalized to (0,1) by the following equation,
%\[
%\overline{\mathbf X}_i = \frac{\mathbf X_i -\mathbf X^{i}_{min}}{\mathbf X^{i}_{max}-\mathbf X^{i}_{min}}
%\]
%where $\mathbf X_i$ represent the $i$-th slice of tensor $\mathcal X$. $\mathbf X^i_{min}$ and $\mathbf X^i_{max}$ represent the minimum and maximum value in $\mathbf X_i$.
\[
\text{err}_i = \frac{\|\mathbf X_i-\overline{\mathbf X}_i\|_F}{\|\mathbf X_i\|_F}
\]
where $\mathbf X_i$ represents the $i$-th frontal slice of $\mathcal X$ and $\overline{\mathbf X}_i$ represents the reconstructions of $\mathbf X_i$ ($i=1,...,4$). The relative reconstruction errors are shown in TABLE \ref{table1}.
\begin{table}[h]
\renewcommand{\arraystretch}{1.3}
\small
\centering
\begin{tabular}{ccccc}
  \hline
  Methods & Temperature & Humidity&Light  & Voltage\\
  \hline
TBV-DR &0.0164 &0.0064&6.4392e-05&0.0290\\
\hline
CP &0.0381&0.0154&9.2701e-04&0.0316\\
  \hline
\end{tabular}
    \caption{Relative reconstruction errors of TBV-DR and CP on Berkeley dataset.}\label{table1}
\end{table}

\begin{figure*}
\begin{center}    \subfloat{\includegraphics[width=65mm,height=50mm]{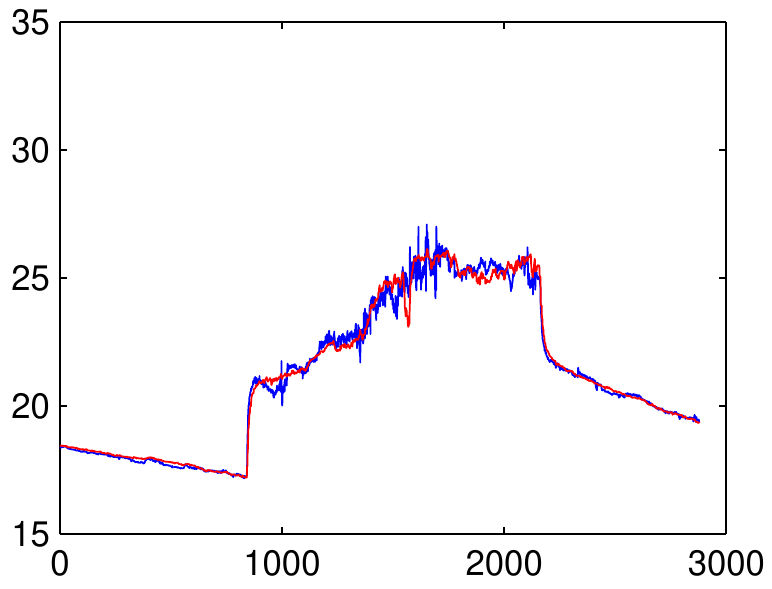}}     \subfloat{\includegraphics[width=65mm,height=50mm]{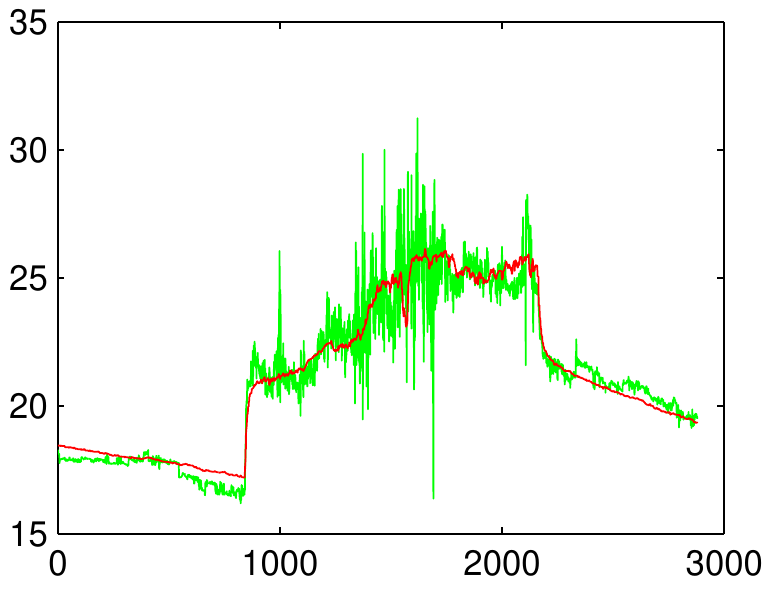}} \\     \subfloat{\includegraphics[width=65mm,height=50mm]{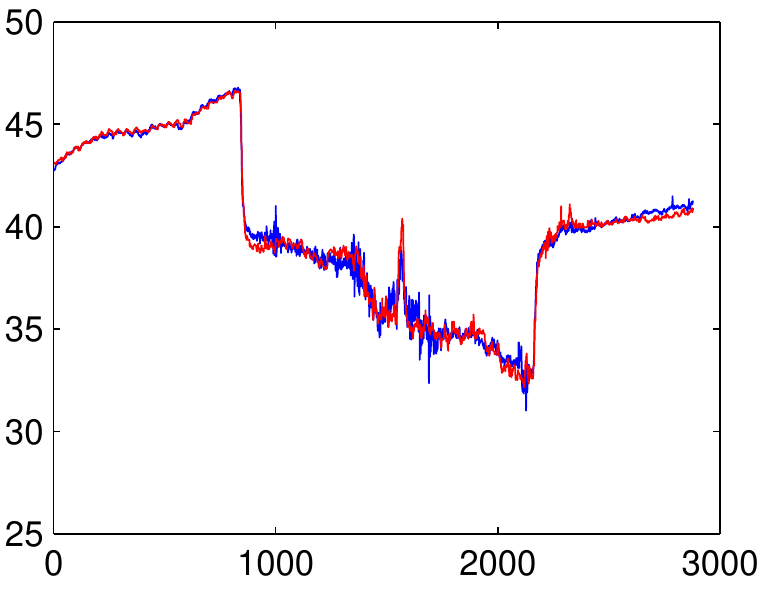}}     \subfloat{\includegraphics[width=65mm,height=50mm]{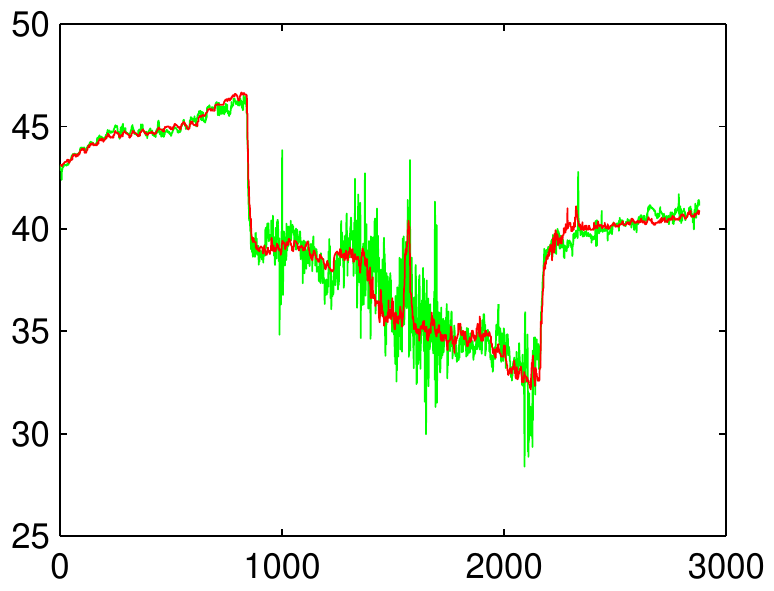}} \\    \subfloat{\includegraphics[width=65mm,height=50mm]{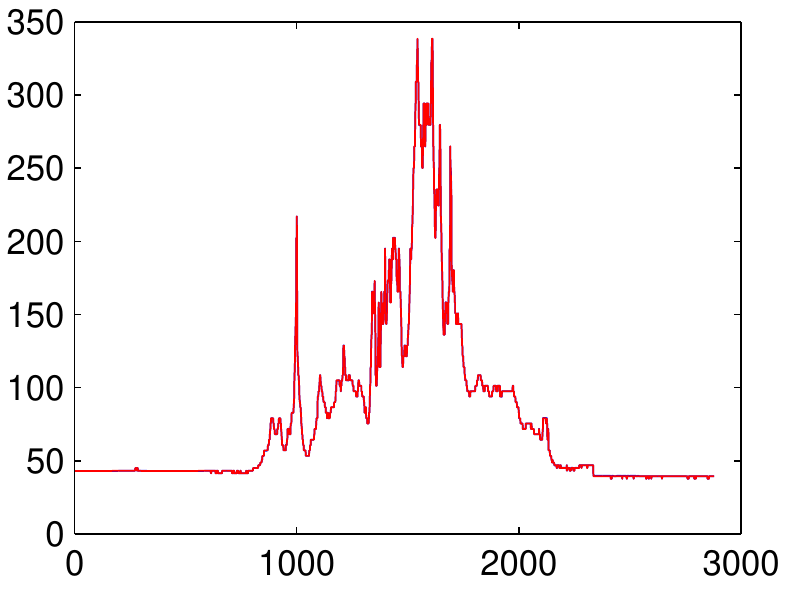}}     \subfloat{\includegraphics[width=65mm,height=50mm]{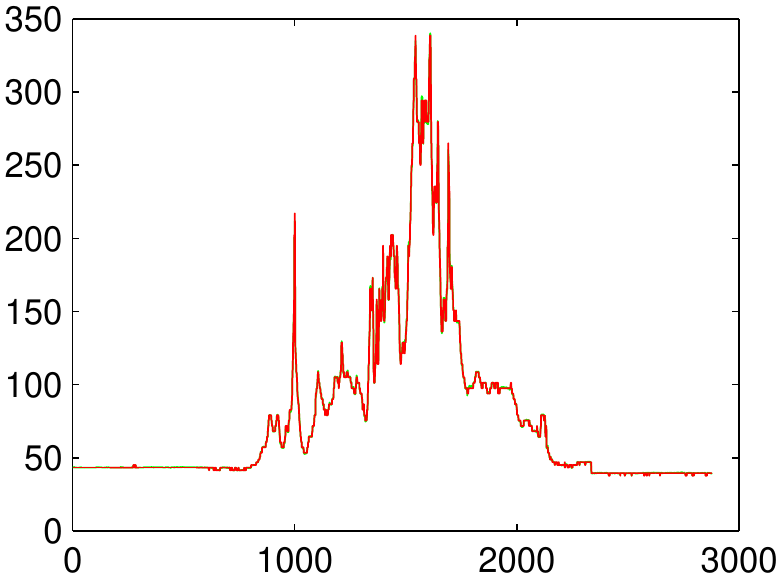}} \\    \subfloat{\includegraphics[width=65mm,height=50mm]{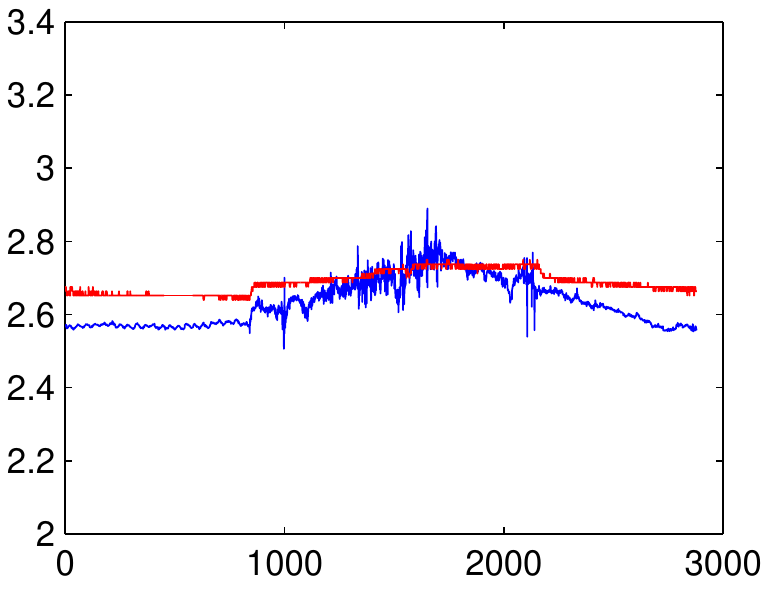}}     \subfloat{\includegraphics[width=65mm,height=50mm]{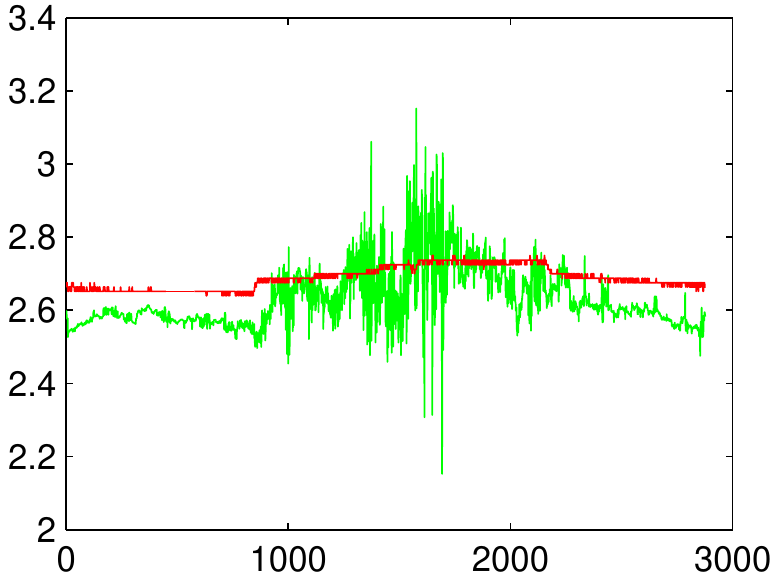}} \\
    \end{center}
    \caption{The reconstruction curves of the first sensor. The red curves represent original data, the blue curves are our results and the green curves are the results of CP. From the first row to the fourth row, the curves are about  temperature, humidity, light and voltage, respectively. }\label{fig5}
\end{figure*}

Fig. \ref{fig5} shows the reconstruction curves of the first sensor. From the first row to fourth row, the curves are about humidity, temperature, light and voltage, respectively. The first column shows the original data (in red) and our results (in blue) while the second column the original data and the reconstruction (in green) achieved by CP, respectively. It seems for the light information both our proposed method and CP can accurately reconstruct the original signal.

From TABLE \ref{table1} and Fig. \ref{fig5} we can conclude that our proposed method can reconstruct original data better than the CP decomposition.

\textbf{IMDB text dataset}: This dataset includes training set and testing set with 25,000 movie reviews. The movie reviews are divided into two categories: negative reviews and positive reviews. We want to test the classification results on IMDB text dataset in this experiment.

In this experiment, we chose 2465 samples as the training set, consisting of 1138 negative reviews and 1327 positive reviews. In addition, we chose 2262 reviews as the testing set, including 1037 negative reviews and 1225 positive reviews. Each selected review contains 180-220 words. If the number of words in a review exceeds 180, then we delete some un-emotional words, like ``a , an, the'' and so on. Thus all reviews are unified to 180 words. We represent each word to a vector with 100 dimension by the word2vector algorithm. Thus each review can be presented as a matrix sample with size of $180\times 100$.
\begin{table}[h]
\renewcommand{\arraystretch}{1.3}
\small
\centering
\begin{tabular}{cc}
  \hline
 Methods & Classification Rate \\
  \hline
PCA &0.7785\\
  \hline
  mixPPCA &0.7882\\
  \hline
GLRAM &0.7887\\
\hline
CP &0.7838\\
  \hline
  Tucker &0.7865\\
  \hline
TBV-DR &0.8232\\
  \hline
\end{tabular}
    \caption{The classification rates on IMDB text dataset.}\label{table2}
\end{table}

The classification results are shown in TABLE \ref{table2}. The reduced dimension is 50 in PCA, mixPPCA, CP and TBV-DR methods. In GLRAM, the reduced dimension is $(r,c) = (8,8)$. This example demonstrates that our proposed method can obtain the better classification rate. It illustrates that our proposed method can perform well on text data.
%-------------------------------------------------------------------------
\section{Conclusions and Future Works}\label{Sec:6}
In this paper, we introduced a probabilistic vectorial dimension reduction model for
high-order tensor data corrupted with Gaussian noises. The model represents a tensor as a linear combination of some basic tensors to achieve a new vectorial representation of tensor for the purpose of dimension reduction. Because using tensor bases leads to an significant increase in the number of parameters to be estimated, we employ  the CP decomposition to construct tensor bases, i.e., expressing the projection base as a sum of a finite number of rank-one tensors.

It has been demonstrated that the proposed model with much less free parameters to be estimated provides comparable expressive capacity against other existing models. %This induces a dimension reduction way for representing high orders tensors as vectorial data.
%This is due to the use of CP decomposition inside the model, %  dramatically reduces the number of parameters to be estimated, resulting a much faster dimension reduction algorithm.
All the key parameters in the probabilistic model can be learned by the variational EM algorithm. Several experiments were conducted to assess the performance of the new model in dimension reduction and feature extraction capability.

In this model, we suppose $\mathcal W$ can be factorized into a sum of component rank-one tensors and the number of components $R$ is given in advance. In fact, estimating the adequate number of components is an important yet difficult problem in multi-way modeling. There are some related works, such as \cite{XiongChenHuangSchneiderCarbonell2010, MorupHansen2009} and \cite{ZhaoZhangCichocki2015}. Motivated by \cite{MorupHansen2009}, we will continue to seek an automatic model selection strategy that can infer the CP rank by introducing a full Bayesian algorithm. To achieve this, we attempt to minimize the dimensionality of latent space, which corresponds to column-wise sparsity of factor matrices. Hence, we employ a sparsity inducing prior over factor matrices by associating an individual hyperparameter with each latent dimension.
It is a very novel question which can be explored in future work.
\section*{Acknowledgments}
This research was supported by National Natural Science Foundation of China (Grant No.61370119, 61390510, 61672071), and also supported by Beijing Natural Science Foundation (Grant No.4172003, 4162010) and Australian Research Council Discovery Projects funding scheme (Project No. DP140102270).

\ifCLASSOPTIONcaptionsoff
  \newpage
\fi
\bibliographystyle{plain}
\bibliography{reference}
\begin{IEEEbiography}[{\includegraphics[width=1in,height=1.25in,clip,keepaspectratio]{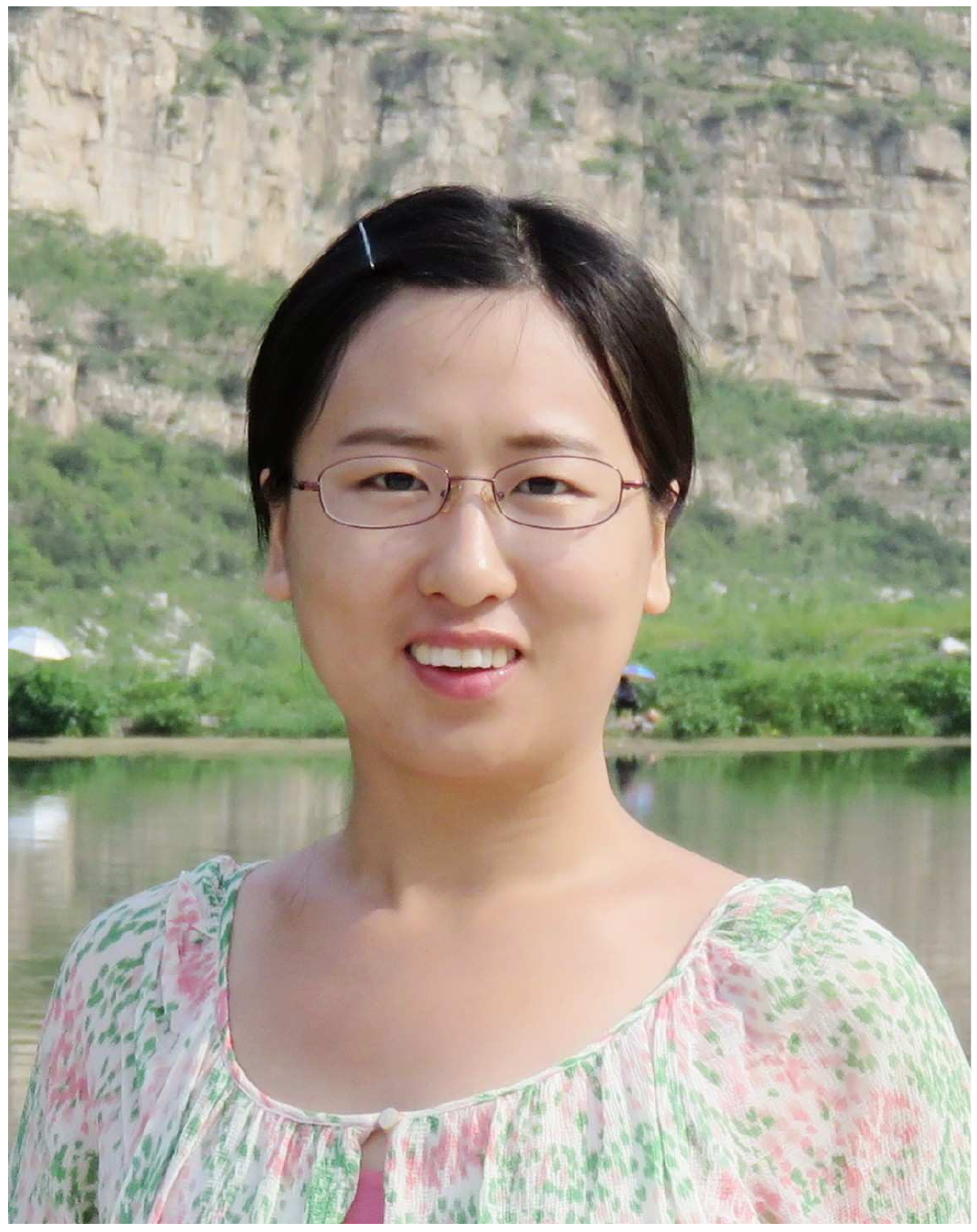}}]{Fujiao Ju} is a PhD student in Computer Science and Technology at the Beijing Key Laboratory of Multimedia and Intelligent Software Technology, Beijing University of Technology, China. She received her Master degree from Beijing University of Technology in Computation Mathematics in 2013. Her research area is Bayesian learning and inference, image analysis.
\end{IEEEbiography}

\begin{IEEEbiography}[{\includegraphics[width=1in,height=1.25in,clip,keepaspectratio]{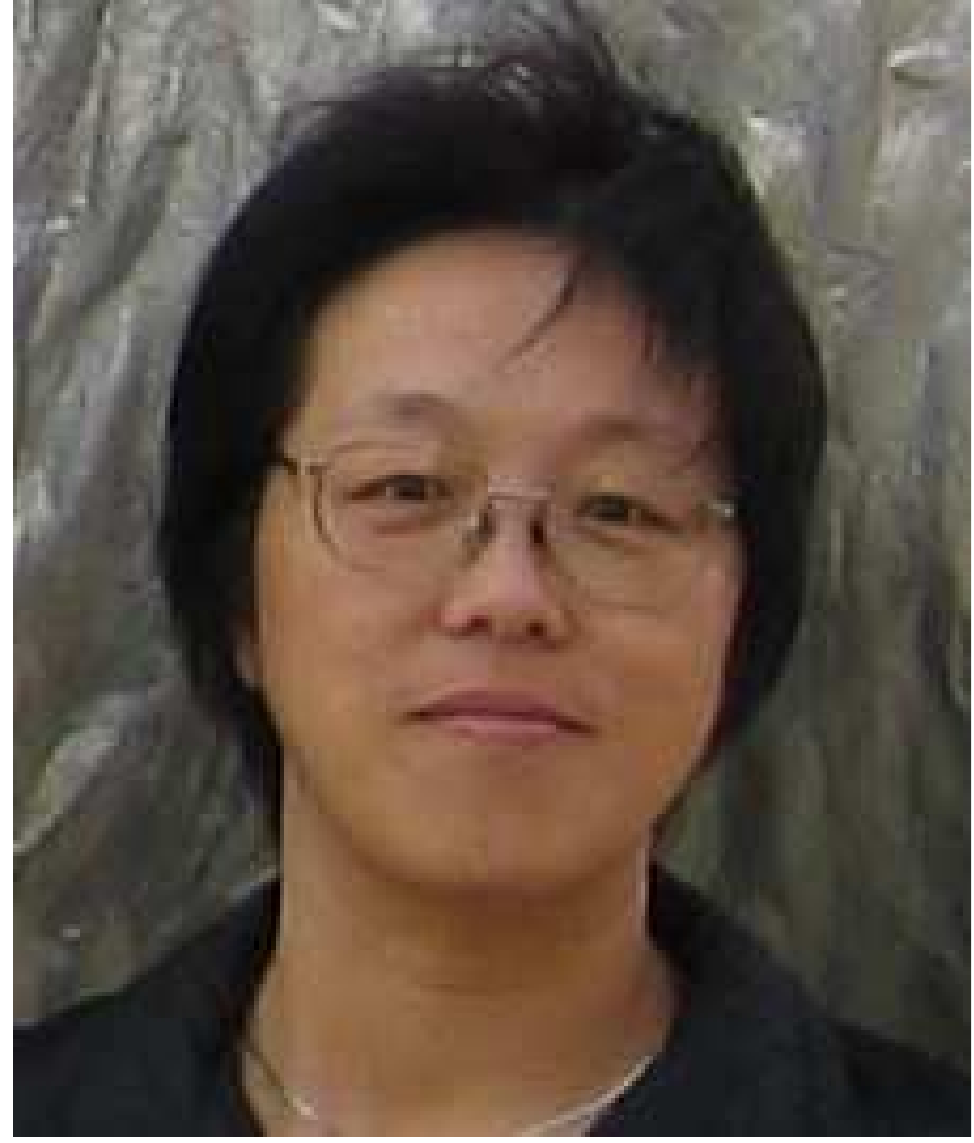}}]{Yanfeng Sun} is a researcher at Beijing Advanced Innovation Center for Future Internet Technology in Computer Science and Technology. She is also a professor at the Beijing Key Laboratory of Multimedia and Intelligent Software Technology, Beijing University of Technology, China. She received her PhD degree from Dalian University of Technology, China in 1993. Her main research interests cover pattern recognition, machine learning and image analysis.
\end{IEEEbiography}

\begin{IEEEbiography}[{\includegraphics[width=1in,height=1.25in,clip,keepaspectratio]{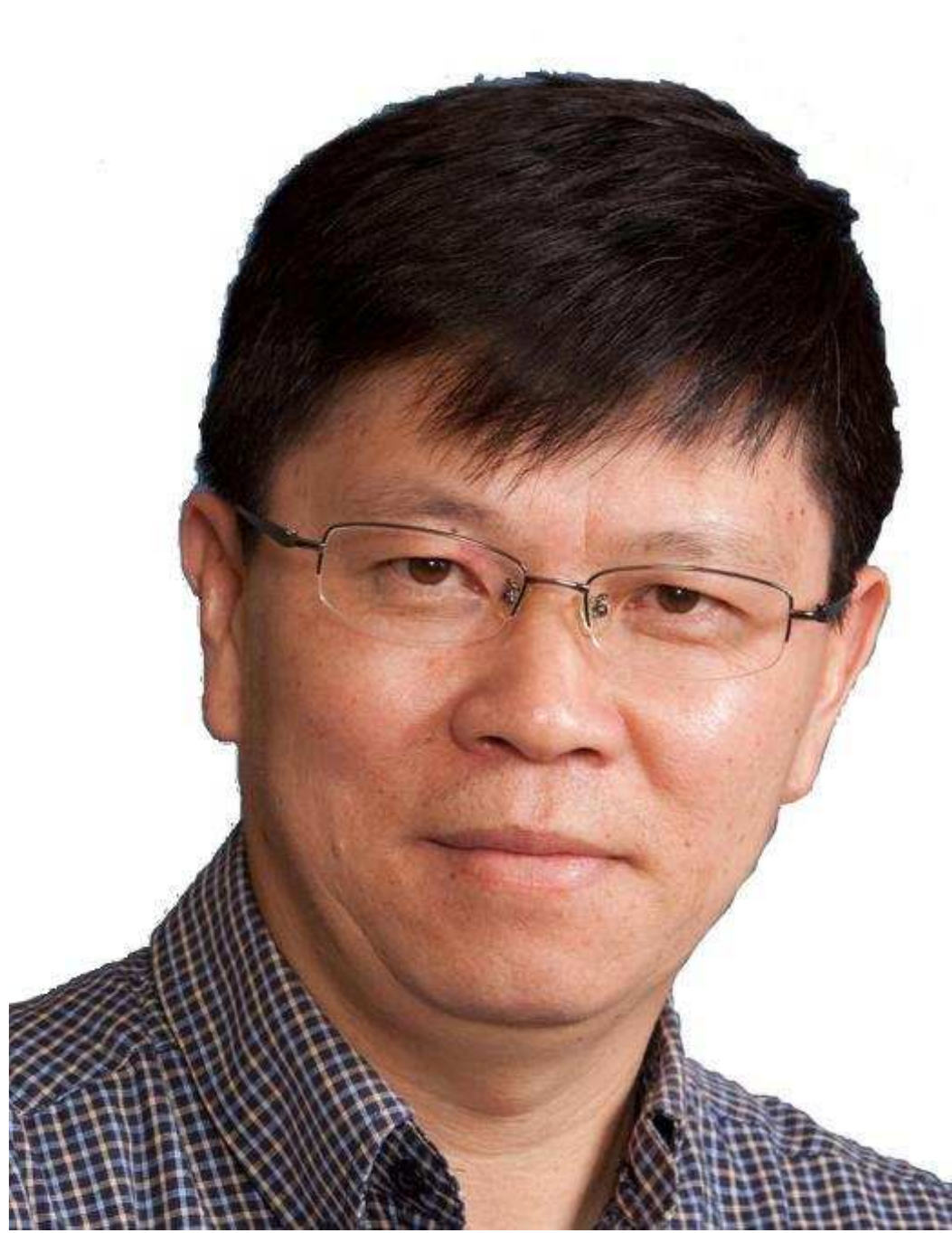}}]{Junbin Gao} graduated from Huazhong University of Science and Technology (HUST),
China in 1982 with BSc. degree in Computational Mathematics and
obtained PhD from Dalian University of Technology, China in 1991. He is a Professor of Big Data Analytics in the University of Sydney Business School at the University of Sydney and was a Professor in Computer Science
in the School of Computing and Mathematics at Charles Sturt
University, Australia. He was a senior lecturer, a lecturer in Computer Science from 2001 to 2005 at
University of New England, Australia. From 1982 to 2001 he was an
associate lecturer, lecturer, associate professor and professor in
Department of Mathematics at HUST. His main research interests
include machine learning, data analytics, Bayesian learning and
inference, and image analysis.
\end{IEEEbiography}

\begin{IEEEbiography}[{\includegraphics[width=1.25in,height=1.25in,clip,keepaspectratio]{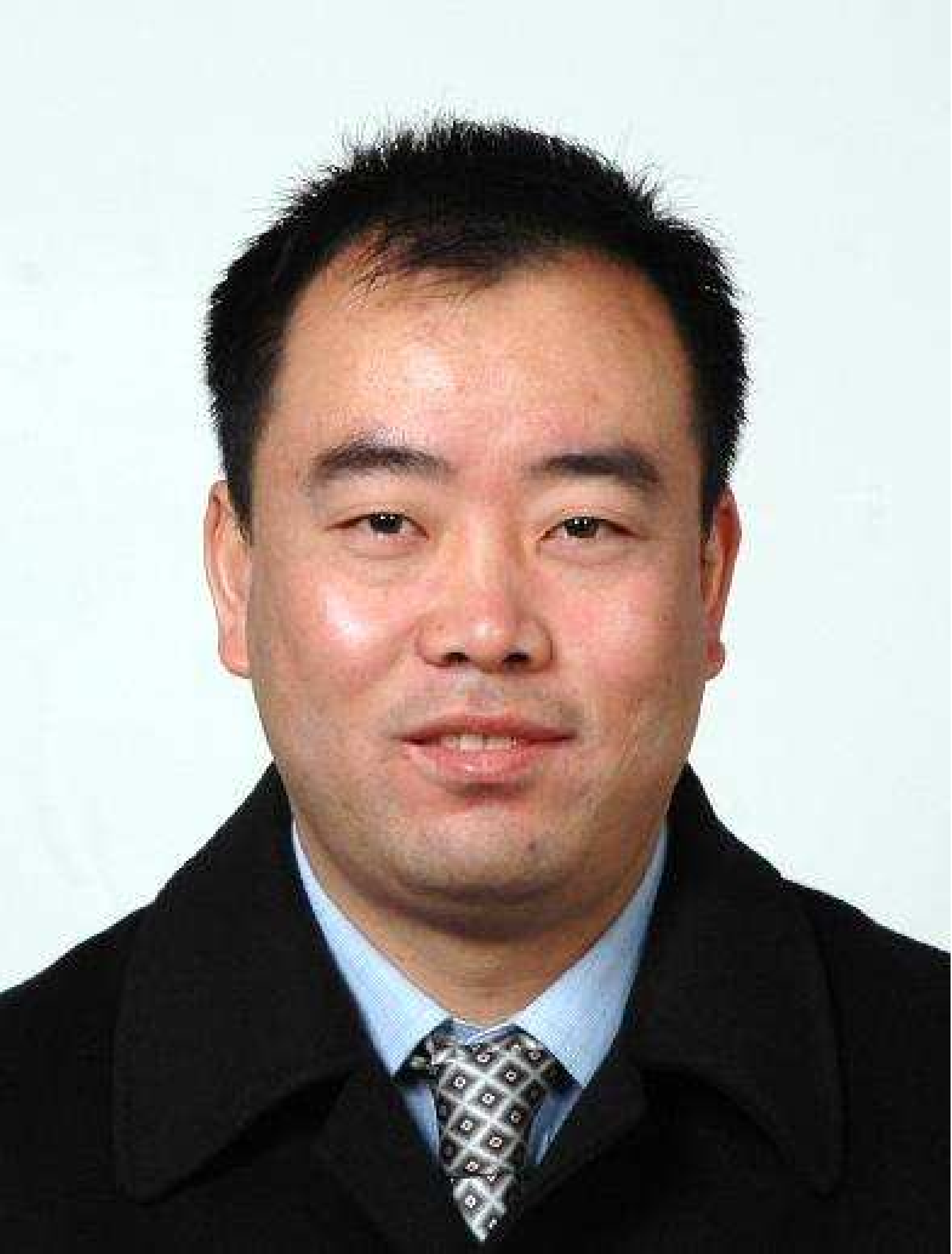}}]{Yongli Hu} received
his PhD degree from Beijing University of Technology in 2005. He is a professor of Computer Science and Technology at the Beijing Key Laboratory of Multimedia and Intelligent Software Technology, Beijing University of Technology, China. He is also a researcher at Beijing Advanced Innovation Center for Future Internet Technology in Computer Science and Technology. His research interests include computer graphics, pattern recognition and multimedia technology.
\end{IEEEbiography}

\begin{IEEEbiography}[{\includegraphics[width=1.4in,height=1.25in,clip,keepaspectratio]{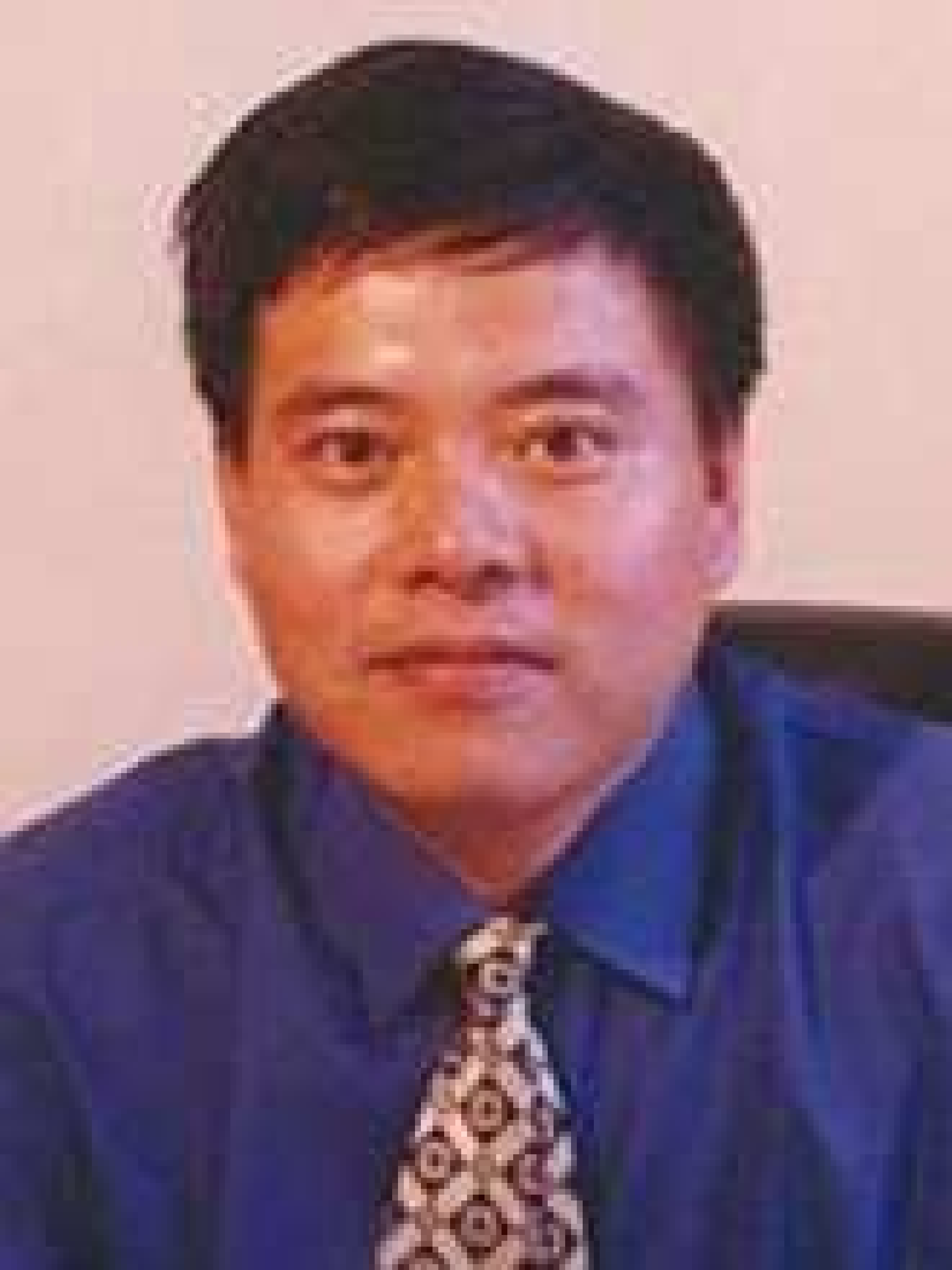}}]{Baocai Yin}  received his PhD from Dalian University of Technology in 1993. He is a professor of Computer Science and Technology  with the Faculty of Electronic Information and Electrical Engineering, Dalian University of Technology, China. He is a member of the China Computer Federation. His research interests cover multimedia, multifunctional perception and virtual reality.
\end{IEEEbiography}
%\vfill
\end{document}